%% file: emnlp2021.tex
\pdfoutput=1
\documentclass[11pt]{article}

\usepackage[]{emnlp2021}

\usepackage{times}
\usepackage{latexsym}

\usepackage[T1]{fontenc}
\usepackage[utf8]{inputenc}

\usepackage{microtype}

\usepackage{soul}
\usepackage{graphicx}
\usepackage{amsmath}
\usepackage{url}
\usepackage{booktabs}
\usepackage{tabularx}
\usepackage{array,multirow}
\usepackage{subcaption}
\usepackage{float}
\usepackage{makecell}
\usepackage{framed}
\usepackage{enumitem}
\usepackage{titlesec} % Allows creating custom \section's
\usepackage{siunitx}
\usepackage{verbatim}
\usepackage{mathtools}
\usepackage[normalem]{ulem}
\usepackage{lipsum}
\usepackage{dingbat}

\newcommand{\STAB}[1]{\begin{tabular}{@{}c@{}}#1\end{tabular}}
\newcommand{\rotate}[2]{\multirow{#1}{*}{\STAB{\rotatebox[origin=c]{90}{#2}}}}

\definecolor{mygray}{gray}{0.7}

\usepackage{amsfonts, amsthm, bm, amssymb}

\usepackage{listings}
\newcommand{\lex}[1]{\textit{#1}}
\newcolumntype{C}{>{\centering\arraybackslash}X}

\definecolor{royalblue}{HTML}{2058DC}
\definecolor{orangered}{HTML}{E34132}
\definecolor{hlcolor}{rgb}{.9, 1, .4}
\sethlcolor{hlcolor}

\colorlet{shadecolor}{hlcolor}

\newcommand{\hide}[1]{}
\newcommand{\qargu}[3]{\begin{quote}\textbf{Claim:} \lex{#1} \\[1mm] \textbf{#2:} \lex{#3}\end{quote}}
\newcommand{\qargutwo}[2]{\begin{quote}\textbf{\textit{P}:} \lex{#1} \\[1mm] \textbf{\textit{H}:} \lex{#2}\end{quote}}
\newcommand{\RR}{\mathbb{R}}

\title{Knowledge-Enhanced Evidence Retrieval\\for Counterargument Generation}

\author
{
    Yohan Jo$^1$, Haneul Yoo$^2$, JinYeong Bak$^3$, Alice Oh$^2$, Chris Reed$^4$, Eduard Hovy$^1$ \\
    $^1$Carnegie Mellon University, USA,
    $^2$KAIST, South Korea \\
    $^3$Sungkyunkwan University, South Korea, 
    $^4$University of Dundee, UK \\
    \texttt{yohanj@cs.cmu.edu}, \texttt{haneul.yoo@kaist.ac.kr}, \texttt{jy.bak@skku.edu},\\ \texttt{alice.oh@kaist.edu}, \texttt{c.a.reed@dundee.ac.uk}, \texttt{hovy@cmu.edu}
}

\begin{document}

\onecolumn
%\begin{verbatim}
\begin{lstlisting}
@inproceedings{jo-2021-kenli,
    title = "Knowledge-Enhanced Evidence Retrieval for Counterargument Generation",
    author = "Jo, Yohan and Yoo, Haneul and Bak, JinYeong and Oh, Alice and Reed, Chris and Hovy, Eduard",
    booktitle = "Findings of the Association for Computational Linguistics: EMNLP 2021",
    month = nov,
    year = "2021",
    address = "Online",
    publisher = "Association for Computational Linguistics"
}
\end{lstlisting}
%\end{verbatim}

\newpage
%\twocolumns

\maketitle
\begin{abstract}
Finding counterevidence to statements is key to many tasks, including counterargument generation. We build a system that, given a statement, retrieves counterevidence from diverse sources on the Web. At the core of this system is a natural language inference (NLI) model that determines whether a candidate sentence is valid counterevidence or not. Most NLI models to date, however, lack proper reasoning abilities necessary to find counterevidence that involves complex inference. Thus, we present a knowledge-enhanced NLI model that aims to handle causality- and example-based inference by incorporating knowledge graphs. Our NLI model outperforms baselines for NLI tasks, especially for instances that require the targeted inference. In addition, this NLI model further improves the counterevidence retrieval system, notably finding complex counterevidence better.\footnote{Source code and data are available at \url{https://github.com/yohanjo/kenli}.}
\end{abstract}

\section{Introduction\label{sec:intro}}

\input{texes/1_Introduction}

\section{Related Work\label{sec:related_work}}
\input{texes/2_RelatedWork}

\section{Knowledge-Enhanced NLI\label{sec:nli}}
\input{texes/3_NLI}

\section{Retrieval of Counterevidence\label{sec:retrieval}}
\input{texes/4_Retrieval}

\section{Conclusion\label{sec:conclusion}}
\input{texes/5_Conclusion}

\section*{Acknowledgments}
This work was partly supported by CMU's GuSH funding. JinYeong Bak is partly supported by Institute of Information \& communications Technology Planning \& Evaluation (IITP) grant funded by the Korea government (MSIT) (No.2019-0-00421, AI Graduate School Support Program (Sungkyunkwan University)) and partly by the MSIT, Korea, under the High-Potential Individuals Global Training Program (2020-0-01550) supervised by the IITP. Alice Oh and Haneul Yoo are supported by Institute of Information \& communications Technology Planning \& Evaluation (IITP) grant funded by the Korea government (MSIT) (No.2017-0-01779, A machine learning and statistical inference framework for explainable artificial intelligence).

\bibliographystyle{acl_natbib}
\bibliography{custom}
%\bibliography{customclean}

% \newpage
% ~
\newpage
\appendix
\input{texes/Appendix}

\end{document}

%% file: texes/1_Introduction.tex
Generating counterarguments is key to many applications, such as debating systems \cite{Slonim:2018.project.debater}, essay feedback generation \cite{Woods:2017eh}, and legal decision making \cite{Feteris.2017.legal}. 
In NLP, many prior studies have focused on generating counterarguments to the main conclusions of long arguments, usually motions. Although such counterarguments are useful, argumentative dialogue is usually interactive and synchronous, and one often needs to address specific statements in developing argument. For instance, in the ChangeMyView (CMV) subreddit, challengers often quote and counter specific statements in the refuted argument, where 41\% of these attacks are about factual falsehood, such as exceptions, feasibility, and lack of evidence \cite{Jo2020:attackable}.
Hence, the scope of our work is narrower than most prior work. Instead of generating a counterargument to a \textit{complete argument}, we aim to find counterevidence to \textit{specific statements} in an argument. This counterevidence may serve as essential building blocks for developing a larger counterargument and also allow for more interactive development of argumentation.

\begin{figure}[t]
    \centering
    \includegraphics[width=.65\linewidth]{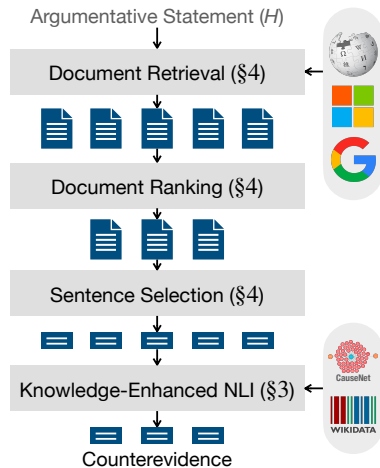}
    \caption{Architecture overview.}
    \label{fig:architecture}
\end{figure}

We adopt a popular fact-verification framework \cite{thorne-etal-2018-fever}: given a statement to refute, we retrieve relevant documents from the Web and select counterevidence (Figure~\ref{fig:architecture}).
At the core of this framework is a module that determines whether a candidate sentence is valid counterevidence to the given statement. A natural choice for this module is a natural language inference (NLI) model. But NLI models to date have shown a lack of reasoning abilities \cite{Williams.2020.anlizing}, which is problematic because counterarguments often involve complex inference. To overcome this limitation, we enhance NLI by focusing on two types of inference informed by argumentation theory \cite{Walton:2008schem}. The first is \textit{argument from examples}, as in:
\begingroup
\addtolength\leftmargini{-5mm}
\qargu{Vegan food reduces the risk of diseases.}{Counterevidence}{Legume protein sources can result in phytohemagglutinin poisoning.}
The inference is based on the fact that ``legume protein sources'' and ``phytohemagglutinin poisoning'' are \textbf{examples} of ``vegan food'' and ``diseases'', respectively. The second type of inference is \textit{argument from cause-to-effect}, as in:
\qargu{Veganism reduces the risk of diabetes.}{Counterevidence}{Vegan diets suffer from poor nutrition.}
The inference is based on the fact that poor nutrition can \textbf{cause} diabetes. 
\endgroup

In order to handle causality- and example-based inference, we develop a knowledge-enhanced NLI model (\S{\ref{sec:nli}}). By incorporating two knowledge graphs---CauseNet and Wikidata---into the model while training on public NLI datasets, the accuracy of the NLI model improves across NLI datasets, especially for challenging instances that require the targeted inference.

We integrate this NLI model into the entire retrieval system to find counterevidence to argumentative statements from two online argument platforms, ChangeMyView (CMV) and Kialo (\S{\ref{sec:retrieval}}). 
We demonstrate that our knowledge-enhanced NLI model improves the system, finding more complex counterevidence. We also conduct in-depth analyses of the utility of different types of source documents and document search methods (Wikipedia, Bing, and Google). 

Our contributions are as follows:
\begin{itemize}[topsep=0pt,itemsep=0pt,parsep=0pt,partopsep=0pt]
    \item A knowledge-enhanced NLI model to handle causality- and example-based inference.
    \item A counterevidence retrieval system improved by the NLI model, along with analyses of different document types and search methods.
    \item A new challenging dataset of counterevidence retrieval, along with all search results and retrieved documents from Bing and Google.
\end{itemize}

%% file: texes/2_RelatedWork.tex
\subsection{Counterargument Generation}
In NLP, there are two main approaches to counterargument generation. \textbf{Retrieval-based} approaches retrieve existing arguments from debates that best serve as counterarguments, based on how similar a claim is to the target argument \cite{Thu:2018.debater} and how dissimilar a premise is \cite{Wachsmuth:2018retr}.
Some studies retrieve texts that contain negative consequences of the target argument \cite{Reisert:2015.toulmin,Sato:2015.retr_cause}. Recently, a human-curated corpus was developed \cite{Orbach.2020.echo}.

\textbf{Neural language generation} approaches take the target argument as input and generate a counterargument using a neural network \cite{Hua:2018.gen,Hua:2019.gen}. These approaches still retrieve evidence sentences from Wikipedia or news articles that are similar to the target argument, which are fed to a neural network to decode a counterargument. 
%The quality of automatically generated counterarguments is yet lower than that of simply concatenating evidence passages, in terms of topical relevance and oppositeness. 
%The quality of evidence passages was not examined in detail. Hence, 
Our work is complementary to these studies, as high-quality counterevidence is essential to decoding high-quality counterarguments.

Most of these studies build a counterargument against an \textit{entire} argument. Thus, generated counterarguments might counter the main conclusion of the target argument without addressing specific points in it. In contrast, our work aims to find counterevidence that directly addresses specific statements in the target argument.

\subsection{Fact Verification}
Since we want to find counterevidence to specific statements in the target argument, our work is closely related to fact verification \cite{Li:2020.fake_news.fever}. 
Recently, this research area has garnered much attention, especially with the emergence of the FEVER (Fact Extraction and VERification) task \cite{thorne-etal-2018-fever}. The FEVER task aims to predict the veracity of statements, and most approaches follow three steps: document retrieval, sentence selection, and claim verification. 
%While early studies focused mostly on evidence representations \cite{Ma:2019.fever.hier.attn,Tokala.2019}, later 
Recent studies examined homogeneous model architectures across different steps \cite{Tokala.2019,Nie.2019.fnli,Nie:2020.compound.fever}. 
Especially BERT has been shown to be effective in both retrieval and verification \cite{Soleimani:2020.bert.fever}, and a joint model of BERT and pointer net achieved state-of-the-art performance in this task \cite{Hidey.2020.deseption}. Our work builds on this model (\S{\ref{sec:retrieval}}).

\subsection{Knowledge-Enhanced Language Models}
The last step of fact verification, i.e., claim verification, relies heavily on natural language inference (NLI) between an evidence text and a statement to verify. Recently, transformer-based language models (LMs) have been prevailing in NLI \cite{Nie.2020anli}, but they still show a lack of reasoning abilities \cite{Williams.2020.anlizing}. Hence, researchers have tried to improve LMs by integrating knowledge, mainly via two approaches.

The first is to exploit knowledge graphs (KGs) mainly to learn better embeddings of tokens and entities \cite{Wang:2019.kepler,Peters.2019.knowbert,Zhang.2019ernie,Lauscher.2020.adapter}. Once learning is done, the model does not require external knowledge during inference. 
The second type of models use the triple information of entities linked to the input text during inference, either by encoding knowledge using a separate network \cite{He.2020.bert-mk,Chen.2018kim} or by converting the KG to input text tokens \cite{Liu:2020.kbert}.
Our work adopts the second approach, as it depends less on the snapshot of the KG used for pretraining. But our model design has clear distinctions from previous work in the way that KGs are integrated and entity paths are taken.

%% file: texes/3_NLI.tex
A natural language inference (NLI) model is the core of our entire system (Figure~\ref{fig:architecture}). Given a statement to refute, the system retrieves and ranks relevant documents, and then obtains a set of candidate sentences for counterevidence. For each candidate, the NLI model decides whether it entails, contradicts, or neither the statement. 
In this section, we first motivate the model design and explain our Knowledge-Enhanced NLI model (KENLI), followed by evaluation settings and results.

\subsection{Motivation}
Many NLI models have difficulty in capturing the relation between statements when their words are semantically far apart.
For instance, if a statement refutes another based on example- or causality-based inference using technical terms (e.g., legume protein sources as an example of vegan food), the semantic gap between the words can make it hard to capture the relation between the two statements without explicit knowledge.

To reduce semantic gaps between words, our method aims to bridge entities in the two statements using a knowledge graph (KG) so that the information of an entity in one statement flows to an entity in the other statement, along with the information of the intermediate entities and relations on the KG. This information updates the embeddings of the tokens linked to the entities.

\subsection{Model}
\begin{figure*}[th]
    \centering
    \includegraphics[width=.95\linewidth]{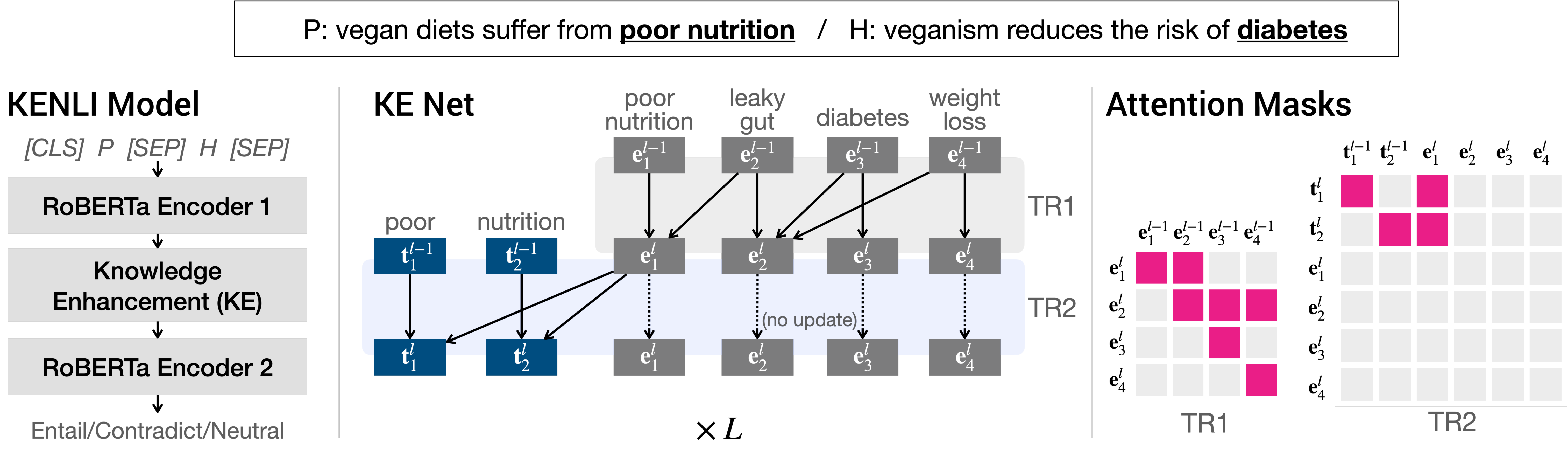}
    \caption{KENLI Model. This example illustrates using two KG paths: ``poor nutrition $\xrightarrow{\text{cause}}$ leaky gut $\xrightarrow{\text{cause}}$ diabetes'' and ``poor nutrition $\xrightarrow{\text{cause}}$ leaky gut $\xrightarrow{\text{cause}}$ weight loss''.}
    \label{fig:ke-nli}
\end{figure*}

KENLI (Figure~\ref{fig:ke-nli} left) is based on RoBERTa-base \cite{liu2019roberta}, which takes a pair of premise $P$ and hypothesis $H$ as input and computes the probability of whether their relation is entailment, contradiction, or neutral. To bridge entities between $P$ and $H$, the \textbf{Knowledge Enhancement (KE) Net} is inserted between two layers (e.g., 10th and 11th layers), splitting RoBERTa into \textbf{Encoder1} and \textbf{Encoder2}. It updates intermediate token embeddings from Encoder1 and feeds them to Encoder2. The final prediction is made through a fully-connected layer on top of the CLS embedding.

The KE Net (Figure~\ref{fig:ke-nli} middle) exploits a knowledge graph (KG) where nodes are entities and edges are directed relations between entities (e.g., `instance\_of', `cause'). Its main goal is to let information flow between entities in $P$ and $H$ through the KG. Suppose the KG has a set of relations $R = \{r_i\}_{i=1}^{|R|}$. 
For each input text pair, $T = \{t_i\}_{i=1}^{|T|}$ is the tokens in $P$ that are linked to entities. Their initial embeddings $\{\mathbf{t}^0_i\}_{i=1}^{|T|}$ are the intermediate token embeddings from Encoder1.
$E = \{e_i\}_{i=1}^{|E|}$ denotes entities under consideration, with initial embeddings $\{\mathbf{e}^0_i\}_{i=1}^{|E|}$. 
Considering all entities in the KG for every input pair is computationally too expensive. Recall that our motivation is to bridge entities between $P$ and $H$. Hence, for each input pair, we first include entity paths whose source is in $P$ and destination is in $H$. We add more destinations with the constraint that the total number of considered entities is no greater than $\lambda$ and the length of each path is no greater than $\nu$ ($\lambda$ and $\nu$ are hyperparameters).
To obtain $\mathbf{e}^0_i$, we simply encode the name of each entity with RoBERTa Encoder1 and sum all the token embeddings. 
%Note that one entity may be linked to multiple tokens if its name consists of multiple tokens or it occurs multiple times in $P$.

The KE Net is a stack of KE cells. Each KE cell handles one-hop inference on the KG using two transformers TR1 and TR2. TR1 updates each entity embedding based on its neighboring entities, and TR2 updates token embeddings based on the embeddings of linked entities. More specifically, in the $l$-th KE cell, TR1 takes $\{\mathbf{e}^{l-1}_i\}_{i=1}^{|E|}$ as input and updates their embeddings using self-attention. Each attention head corresponds to each relation, and the attention mask for the $k$-th head $M^k \in \RR^{|E| \times |E|}$ allows information flow between entities that have the $k$-th relation:
\begin{align*}
    M^k_{ij} = \left\{ \begin{array}{ll}
        1 & \text{if $i=j$ or $(e_i, r_k, e_j) \in $ KG} \\
        0 & \text{otherwise}.
    \end{array}\right.
\end{align*}
TR2 takes the concatenation of $\{\mathbf{t}^{l-1}_i\}_{i=1}^{|T|}$ and $\{\mathbf{e}^l_i\}_{i=1}^{|E|}$ as input and updates the token embeddings using one attention head with attention mask $M \in \RR^{|T+E| \times |T+E|}$:
\begin{align*}
    M_{ij} = \left\{ \begin{array}{ll}
        1 & \text{if $i \leq |T|$ and } \\
          & \text{~~~($i=j$ or $t_i$ is linked to $e_{j-|T|}$)} \\
        0 & \text{otherwise}.
    \end{array}\right.
\end{align*}
Entity embeddings are not updated in TR2.

After token embeddings are updated by $L$ KE cells (i.e., $L$-hop inference), the token embedding of $t_i$ is updated as
$\mathbf{t}_i \leftarrow \mathbf{t}^0_i + \mathbf{t}^L_i$
and fed to Encoder2 along with the other token embeddings in the input.

\subsection{Knowledge Graphs}
Our work uses two knowledge graphs: CauseNet and Wikidata. \textbf{CauseNet} \cite{Heindorf.2020.causenet} specifies \textit{claimed} causal relations between entities, extracted from Wikipedia and ClueWeb12 based on linguistic markers of causality (e.g., ``cause'', ``lead'') and infoboxes. 
We discard entity pairs that were identified by less than 5 unique patterns, since many of them are unreliable. This results in total 10,710 triples, all having the `cause' relation.

\textbf{Wikidata} \cite{Vrandecic.2014.wikidata} is a database that specifies a wide range of relations between entities. We use the October 2020 dump and retain triples that have 8 example-related relations: instance\_of, subclass\_of, part\_of, has\_part, part\_of\_the\_series, located\_in\_the\_administrative\_territorial\_entity, contains\_administrative\_territorial\_entity, and location. The importance of information about physical and temporal containment in NLI was discussed recently \cite{Williams.2020.anlizing}. This filtering results in 95M triples, which we call \textbf{WikidataEx}.

\subsection{Data}
Our data mainly come from public NLI datasets: MNLI \cite{Williams:2018mnli}, ANLI \cite{Nie.2020anli}, SNLI \cite{Bowman.2015snli}, and FEVER-NLI\hide{\footnote{FEVER-NLI is an adaptation of the FEVER dataset to the NLI format. Each claim is paired with an evidence statement. Each pair is labeled as entailment, contradiction, or neither, which correspond to the original labels of support, refute, and not-enough-info, respectively.}} \cite{Nie.2019.fnli}. We split the data into train, validation, and test sets as originally or conventionally set up for each dataset (Table~\ref{tab:datasets}). Due to limited computational resources, our training set includes only MNLI and ANLI.

The public NLI datasets alone may not include enough instances that require example- and causality-based inference. As a result, the NLI model may not learn to exploit the KGs well. To alleviate this issue, we generate synthetic NLI pairs that are built on example-based inference as follows (details are in Appendix~\ref{sec:annot_example}). Given a pair of $P$ and $H$ in the public datasets, we modify $P$ to $P'$ by replacing an entity that occurs in both $P$ and $H$ with an incoming entity on WikidataEx (e.g., ``England'' with ``Yorkshire''). This achieves two goals. First, $P'$ includes an entity that is an example of another entity in $H$ so that the $(P',H)$ pair requires example-based inference, with the same expected relation as the $(P,H)$ pair. Second, this example relation comes from our KG so that the NLI model learns how to use the KG.
%To avoid broken or implausible sentences, we retain $P'$ only if its perplexity is lower than that of $P$ based on GPT2. 
Generating similar NLI pairs for causality-based inference is more challenging, and we leave it to future work. 

\begin{table}[t]
    \centering
    \small
    \begin{tabularx}{.8\linewidth}{Xrrr}\toprule
        Dataset & Train & Val & Test \\
        \midrule
        % MNLI & 392,702 & 9,815 & $^\dagger$9,796 \\
        % MNLI-MM & -- & 9,832 & $^\dagger$9,847 \\
        MNLI & 392,702 & -- & 9,815 \\
        MNLI-MM & -- & -- & 9,832 \\
        ANLI & 162,865 & 3,200 & 3,200 \\
        SNLI & \hide{549,367}-- & 9,842 & 9,824 \\
        SNLI-Hard & -- & -- & 3,261 \\
        FEVER-NLI & \hide{208,346}-- & 9,999 & 9,999 \\
        \midrule
        Example-NLI & 30,133 & 2,867 & 3,468 \\
        ANLI-Contain & -- & -- & 277 \\
        ANLI-Cause & -- & -- & 1,078 \\
        BECauSE & -- & -- & 2,814 \\
        \bottomrule
    \end{tabularx}
    \caption{Number of NLI pairs by dataset.}
    \label{tab:datasets}
\end{table}

\paragraph{Inference Evaluation:}
We use additional datasets to evaluate NLI models' inference abilities. For example-based inference, we first use a diagnostic subset of ANLI that has been annotated with various categories of required inference, such as counting, negation, and coreference \cite{Williams.2020.anlizing}. We choose the instances of the `Containment' category, which requires inference on part-whole and temporal containment between entities (\textbf{ANLI-Contain}). 
In addition, we use the test set of our \textbf{Example-NLI} data after manually inspecting their labels.

For causality-based inference, we use the instances in the diagnostic ANLI set that belong to the `CauseEffect' and `Plausibility' categories (\textbf{ANLI-Cause}). They require inference on logical conclusions and the plausibility of events. 
In addition, we use \textbf{BECauSE 2.0} \cite{Dunietz17:because}, which specifies the `Cause' and `Obstruct' relations between text spans based on linguistic markers of causality. Since it has only two classes, we randomly pair up text spans to generate `neutral' pairs. For reliability, we discard pairs where at least one text comprises only one word. Although this data is not for NLI, we expect that the better NLI models handle the causality between events, the better they may distinguish between the cause, obstruct, and neutral relations. 
See Table~\ref{tab:datasets} for statistics.

\subsection{Experiment Settings}
For KENLI, the KE Net is inserted between the 10th and 11th layers of RoBERTa, although the location of insertion has little effect on NLI performance. The KE Net has a stack of two KE cells, allowing for 2-hop inference on a KG. We test KENLI with CauseNet (\textbf{KENLI+C}) and with WikiedataEx (\textbf{KENLI+E}); we do not combine them so we can understand the utility of each KG more clearly. The maximum number of entities for each input ($\lambda$) and the maximum length of each KG path ($\nu$) are set to 20 and 2, respectively.
To see the benefit of pretraining the KE Net (as opposed to random initialization) prior to downstream tasks, we also explore pretraining it with masked language modeling on the training pairs while the original RoBERTa weights are fixed (\textbf{KENLI+E+Pt} and \textbf{KENLI+C+Pt}). The Adam optimizer is used with a learning rate of 1e-5. See Appendix~\ref{app:reproducibility_kenli} for a reproducibility checklist.

We compare KENLI with three baselines. The first two are state-of-the-art language models enhanced with knowledge graphs. 
\textbf{K-BERT} \cite{Liu:2020.kbert} exploits a KG during both training and inference, by verbalizing subgraphs around the entities linked to the input and combining the verbalized text into the input. 
\textbf{AdaptBERT} \cite{Lauscher.2020.adapter} uses a KG to enhance BERT using bottleneck adapters \cite{Houlsby.2019.adapter}; after that, it is fine-tuned for downstream tasks like normal BERT. We pretrain AdaptBERT for masked language modeling on sentences that verbalize CauseNet (10K) and a subset of WikidataEx (10M) for four epochs. We use the hyperparameter values as suggested in the papers.
The last baseline is \textbf{RoBERTa}-base fine-tuned on the NLI datasets. RoBERTa trained with the ANLI dataset recently achieved a state-of-the-art performance for NLI \cite{Nie.2020anli}.

Input texts are linked to WikidataEx entities by the Spacy Entity Linker\footnote{\url{https://pypi.org/project/spacy-entity-linker/}}.
CauseNet has no public entity linker, so we first stem all entities and input words using Porter Stemmer and then use exact stem matching for entity linking. The stemming allows verbs in input texts to be linked to entities (e.g., ``infected--infection'', ``smokes--smoking'').

\subsection{Results}
\begin{table*}[t]
    \centering
    \small
    \begin{tabularx}{\linewidth}{@{} lXXXXXXX p{1mm} XXXX}\toprule
         & \multicolumn{7}{c}{NLI Evaluation} & & \multicolumn{4}{c}{Inference Evaluation} \\
         \cmidrule{2-8} \cmidrule{10-13} 
         & MNLI & MNLI-MM & ANLI & SNLI & SNLI-Hard & FEVER-NLI & Micro Avg &  & Example-NLI & ANLI-Contain & ANLI-Cause & BECau-SE \\
        \midrule
        AdaptBERT+C & 83.0 & 83.6 & 44.7 & 78.8 & 68.2 & 68.2 & 75.4 &  & 58.4 & 42.3 & 35.0 & 27.6 \\
        AdaptBERT+E & 83.2 & 83.5 & 44.7 & 78.7 & 68.4 & 67.8 & 75.4 &  & 58.8 & 43.4 & 34.7 & 27.5 \\
        K-BERT+C & 83.7 & 83.9 & 45.2 & 80.0 & 70.3 & 68.9 & 76.2 &  & 58.9 & 42.4 & 34.7 & 27.2 \\
        K-BERT+E & 83.4 & 83.7 & 46.0 & 79.3 & 69.5 & 69.2 & 76.0 &  & 59.0 & 44.2 & \textbf{35.8} & 26.9 \\
        RoBERTa & 87.3 & 87.0 & \underline{48.6} & 84.2 & 74.6 & 71.9 & 79.7 &  & 61.8 & 47.7 & 35.0 & 27.6 \\
        \midrule
        KENLI+C & \underline{87.5} & 87.1 & 48.2 & \underline{84.3} & 74.8 & 71.4 & 79.7 &  & \underline{62.0} & 48.2 & 35.1 & \underline{27.9} \\
        KENLI+C+Pt & 87.3 & 86.9 & \textbf{48.8} & 84.2 & 74.2 & 71.9 & 79.7 &  & 61.7 & \underline{48.4} & 35.2 & 27.8 \\
        KENLI+E & 87.3 & \textbf{87.2} & 48.5 & 84.2 & \textbf{75.1} & \textbf{72.5}$^\star$ & \underline{79.9}$^\star$ &  & 61.9 & \textbf{49.2} & \underline{35.5} & \textbf{28.0} \\
        KENLI+E+Pt & \textbf{87.6} & \underline{87.1} & 48.4 & \textbf{84.6} & \underline{75.1} & \underline{72.5}$^\star$ & \textbf{80.0}$^\dagger$ &  & \textbf{62.0} & 46.9 & 35.2 & 27.6 \\
        \bottomrule
    \end{tabularx}
    \caption{F1-scores of NLI models by dataset. Statistical significance was measured by the paired bootstrap against the best baseline ($p < 0.05^\star, 0.01^\dagger$). Bold and underline each indicate top1 and top2 results, respectively.}
    \label{tab:nli_accs}
\end{table*}

Table~\ref{tab:nli_accs} shows the F1-scores of each model averaged over 5 runs with random initialization.

In the NLI evaluation, KENLI (rows 6--9) generally outperforms the baseline models (rows 1--5) across datasets. Especially KENLI with WikidataEx (rows 8--9) performs best overall and notably well for difficult datasets (SNLI-Hard, FEVER-NLI, and ANLI). This suggests that KENLI effectively incorporates example-related knowledge, which benefits prediction of nontrivial relations between statements. KENLI with CauseNet (rows 6--7) slightly underperforms KENLI+E, and its average F1-score across datasets is comparable to RoBERTa (row 5). Without pretraining (row 6), it performs slightly better than RoBERTa overall except for two difficult datasets ANLI and SNLI-Hard. With pretraining (row 7), its performance is best for the most difficult dataset ANLI, but slightly lower than or comparable to RoBERTa for the other datasets. This variance in performance across datasets makes it hard to conclude the benefit of CauseNet in general cases.

However, according to the inference evaluation, KENLI's strength is clearer compared to other models. For example-based inference, KENLI+E (row 8) significantly outperforms the other models (ANLI-Contain) or performs comparably well (Example-NLI). Its performance is best or second-best for causality-based inference as well (ANLI-Cause and BECauSE). This suggests that the benefit of example-related knowledge is not limited to example-based inference only. Although KENLI+C (rows 6--7) shows comparable performance to RoBERTa for the general NLI tasks, it consistently outperforms RoBERTa when example- and causality-based inference is required. Example NLI pairs that are classified by only one model are shown in Table~\ref{tab:nli_examples} in Appendix~\ref{app:nli_examples}.

Pretraining KENLI (rows 7 \& 9) does not show a conclusive benefit compared to no pretraining (rows 6 \& 8). Particularly for difficult datasets and inference evaluation, KENLI+E without pretraining (row 8) performs better than pretraining (row 9). The benefit of pretraining for KENLI+C varies depending on the dataset and inference task, making no substantial difference overall.

%% file: texes/4_Retrieval.tex
Our system for counterevidence retrieval builds on DeSePtion \cite{Hidey.2020.deseption}, a state-of-the-art system for the fact extraction and verification (FEVER) task \cite{thorne-etal-2018-fever}. As Figure~\ref{fig:architecture} shows, given a statement to verify, it retrieves and ranks relevant documents, ranks candidate evidence sentences, and predicts whether the statement is supported, refuted, or neither. We adapt DeSePtion to suit our task, where the main contribution is to strengthen the last stage via our knowledge-enhanced NLI model (a detailed comparison between our system and DeSePtion is in Appendix~\ref{sec:deseption}).
We first explain individual stages and then describe evaluation settings and results. 

\subsection{Stages}
\paragraph{Document Retrieval:} 
Documents that may contain counterevidence are retrieved. Given a statement to verify, we retrieve candidate documents from Wikipedia, Bing, and Google. 
For Wikipedia, we use the Spacy Entity Linker to retrieve the articles of Wikidata entities linked to the statement. And for each linked entity, we additionally sample at most five of their instance entities and the corresponding articles, which potentially include counterexamples to the statement. We retrieve additional Wikipedia pages by using named entities in the statement as queries for the wikipedia library\footnote{\url{https://pypi.org/project/wikipedia/}}. We also conduct TF-IDF search using DrQA \cite{Chen:2017drqa} indexed for the FEVER task.
For Bing and Google, we use their search APIs. Wikipedia pages are excluded from their search results, and PDF files are processed using the pdfminer library.
%The three sources provide documents with somewhat different characteristics, and we will compare their utility in \S{\ref{sec:retrieval_results}}.

\paragraph{Document Ranking:}
Retrieved documents are ranked via DeSePtion with some adaptation.
First, RoBERTa is trained to predict whether each document is relevant or not, on the FEVER data. It takes the concatenation of a document snippet and the statement to verify as input. For documents from Bing and Google, we use search result snippets; for Wikipedia, we obtain snippets by concatenating the title of each Wikipedia page with its sentence that is most similar to the statement based on RoBERTa\hide{\footnote{We add all token embeddings in the last layer of RoBERTa and measure cosine similarity.}}.
The last embedding of the CLS token is used as the embedding of the input document, and a pointer net takes these embeddings of all documents and sequentially outputs pointers to relevant documents.

The number of retrieved documents varies depending on the search method, much fewer for the Google API than Wikipedia and Bing in general. Since this imbalance makes it difficult to compare the utility of the different search methods, we keep the number of candidate documents the same across the methods, by ranking documents from different search methods separately and pruning low-ranked documents of Wikipedia and Bing. As a result, the three methods have the same average number of candidate documents per statement ($\sim$8).

\paragraph{Sentence Selection:}
For each statement to verify, we select the top 200 candidate sentences (in all ranked documents) whose RoBERTa embeddings have the highest cosine similarity to the statement.

\paragraph{Relation Prediction:}
We classify whether each candidate sentence is valid counterevidence to the statement to verify. We simply use an NLI model to compute the probability of contradiction and rank sentences by this score. The reason DeSePtion is not used here is described in Appendix~\ref{sec:deseption}.

See Appendix~\ref{app:reproducibility_retr} for a reproducibility checklist.

\subsection{Data\label{sec:retrieval_data}}
Input statements to our system come from two public argument datasets \cite{Jo2020:attackable} collected from the ChangeMyView (CMV) subreddit and Kialo. 
On CMV, the user posts an argument, and other users attempt to refute it often by attacking specific sentences. Each sentence in an argument becomes our target of counterevidence. 
On Kialo, the user participates in a discussion for a specific topic and makes a statement (1--3 sentences) that either supports or attacks an existing statement in the discussion. We find counterevidence to each statement.
To use our resources more efficiently, we discard CMV sentences or Kialo statements that have no named entities or Wikidata entities, since they often do not have much content to refute. We also run coreference resolution for third-person singular personal pronouns using the neuralcoref 4.0 library\footnote{\url{https://github.com/huggingface/neuralcoref}}. 
We randomly select 94 posts (1,599 sentences) for CMV and 1,161 statements for Kialo for evaluation.

\subsection{Evaluation\label{sec:retrieval_eval}}
We evaluate four NLI models. The first three models are directly from \S{\ref{sec:nli}}. That is, \textbf{RoBERTa} is fine-tuned on the NLI data. \textbf{KENLI+C} and \textbf{KENLI+E} are trained with CauseNet and WikidataEx, respectively, without pretraining. 
The last baseline is \textbf{LogBERT}, a state-of-the-art model for argumentative relation classification \cite{Jo:tacl21}. Given a pair of statements, it predicts whether the first statement supports, attacks, or neither the second statement based on four logical relations between them, namely, textual entailment, sentiment, causal relation, and normative relation\footnote{For implementation, BERT-base is fine-tuned for the four classification tasks and then for argumentative relation classification on the Kialo arguments (both normative and non-normative) in the original paper.}. Since LogBERT captures the support and attack relations beyond textual entailment, this baseline would show whether NLI is sufficient for finding counterevidence.

We collect a ground-truth set of labeled data using MTurk. 
First, for each statement to refute, we include in the ground-truth set the top candidate sentence from each model if the probability of `contradiction' is $\geq 0.5$ (i.e., max four sentences).
As a result, the ground-truth set consists of 4,783 (CMV) and 3,479 (Kialo) candidate sentences; they are challenging candidates because at least one model believes they are valid counterevidence.

Each candidate sentence is scored by two Turkers with regard to how strongly it refutes the statement (very weak=0, weak=1, strong=2, and very strong=3). Each candidate sentence is displayed with the surrounding sentences in the original source document as context, as well as a link to the source document. If a candidate is scored as both very weak and very strong, these scores are considered unreliable, and the candidate is scored by a third Turker.
For each candidate, the mean score $s$ is taken as the ground-truth validity as counterevidence: `valid' if $s \geq 1.5$ and `invalid' otherwise. 
More details are described in Appendix~{\ref{sec:annot_evidence}}.

According to the additional question of whether reading the source document is necessary to make a decision for each candidate, about 40\% of answers and 65\% of candidates required reading source documents. This might indicate that three sentences are insufficient for making robust decisions about counterevidence, but it could also be the case that, since our system checks all documents and filter them by relevance in earlier stages, it would not benefit much from more than three sentences. 

We use four evaluation metrics on the ground-truth set. \textbf{Precision}, \textbf{recall}, and \textbf{F1-score} are computed based on whether the model-predicted probability of contradiction for each candidate is $\geq 0.5$. These metrics, however, make the problem binary classification, missing the nuanced degree of validity for each candidate. Thus, we measure \textbf{Kendall's $\boldsymbol{\tau}$} between mean validity scores from human judgments and each model's probability scores. High $\tau$ indicates a good alignment between the human judgment and the model judgment about the strength of validity of each candidate.

\subsection{Results\label{sec:retrieval_results}}
\begin{table}[t]
    \centering
    \small
    \begin{tabularx}{\linewidth}{@{}p{12mm} XXXp{6mm} XXXp{6mm}}\toprule
         &  \multicolumn{4}{c}{CMV} & \multicolumn{4}{c}{Kialo} \\
        \cmidrule(r){2-5} \cmidrule{6-9}
         & Prec & Recl & F1 & $\tau$ & Prec & Recl & F1 & $\tau$ \\
        \midrule
        RoBERTa & 48.3 & 63.6 & 54.9 & 0.002 & 58.0 & 57.0 & 57.5 & 0.022 \\
        KENLI+C & 48.8 & \underline{65.0} & 55.8 & 0.014 & \underline{59.0} & 62.2$^\ddagger$ & 60.6$^\ddagger$ & 0.038$^\dagger$ \\
        KENLI+E & \underline{48.9} & \textbf{71.3}$^\ddagger$ & \textbf{58.0}$^\ddagger$ & \underline{0.015} & 58.0 & \underline{65.2}$^\ddagger$ & \underline{61.4}$^\ddagger$ & \underline{0.038}$^\dagger$ \\
        LogBERT & \textbf{51.4}$^\dagger$ & 61.8 & \underline{56.1} & \textbf{0.031}$^\star$ & \textbf{60.0} & \textbf{66.2}$^\ddagger$ & \textbf{62.9}$^\ddagger$ & \textbf{0.045}$^\dagger$ \\
        \bottomrule
    \end{tabularx}
    \caption{Accuracy of evidence retrieval. For precision, recall, and F1-score, statistical significance was calculated using the paired bootstrap against RoBERTa; for Kendall's $\tau$, the statistical significance of each correlation value was calculated ($p < 0.05^\star, 0.01^\dagger, 0.001^\ddagger$).}
    \label{tab:evidence_accs}
\end{table}

Table~\ref{tab:evidence_accs} summarizes the accuracy of evidence retrieval. Both KENLI+C (row 2) and KENLI+E (row 3) outperform RoBERTa (row 1) for both CMV and Kialo. 
The motivation behind KENLI was to capture statement pairs that require complex inference, by bridging entities with KGs. As expected, KENLI identifies more instances of contradiction that are missed by RoBERTa, as indicated by its high recall. The recall of KENLI+E is substantially higher than RoBERTa's by 7.7 and 8.3 points for CMV and Kialo, respectively, while its improvement of precision is relatively moderate. KENLI+C has a similar pattern but with a smaller performance gap with RoBERTa.

\begin{figure}[t]
    \centering
    \includegraphics[width=\linewidth]{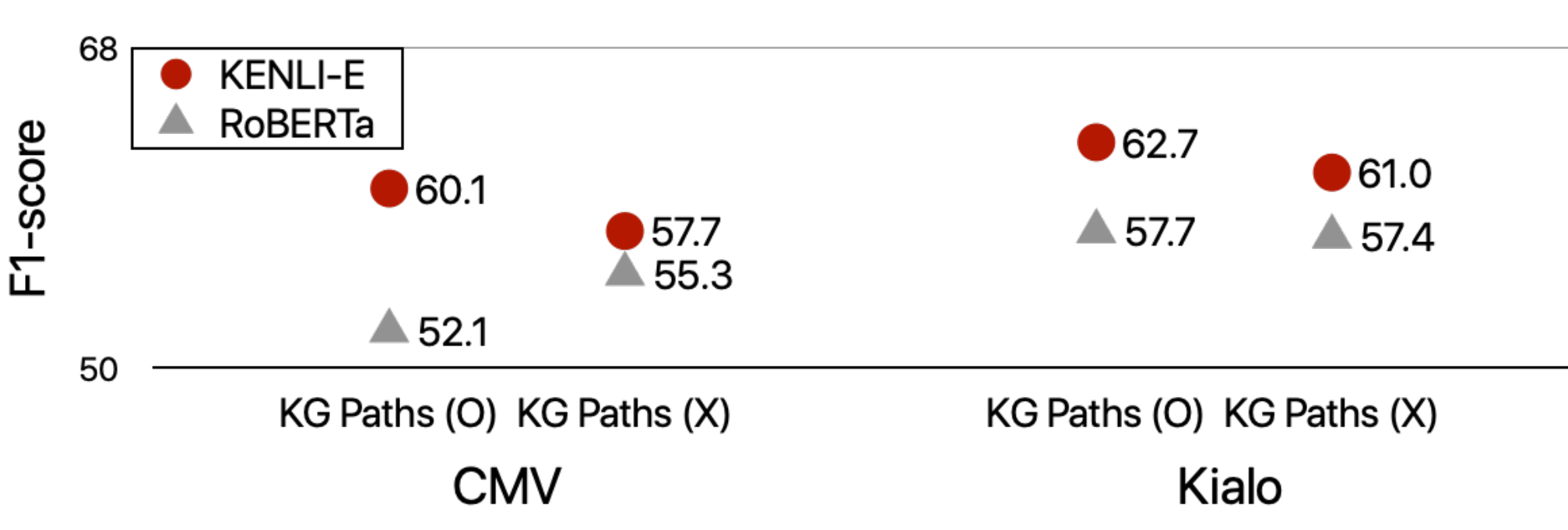}
    \caption{F1-scores of KENLI+E and RoBERTa by the existence of KG paths in candidate sentences.}
    \label{fig:kenli_gap}
\end{figure}

To see if KENLI+E indeed effectively captures counterevidence that requires example-based inference, we broke down its F1-score into one measured on candidate sentences for which KG paths exist between their tokens and the statement's tokens and one measured on the other candidate sentences with no connecting KG paths (Figure~\ref{fig:kenli_gap}). The F1-score gap between KENLI+E and RoBERTa is substantially higher for the candidate sentences where KG paths exist. The gap of recall is even higher, indicating that KENLI+E indeed captures complex counterevidence more effectively than RoBERTa. KG paths that benefit KENLI+E the most include ``player PART\_OF game'', ``Tel Aviv District LOCATED\_IN Israel'', and ``neurovascular system HAS\_PART brain''. 
%That is, KENLI+E and RoBERTa show the largest accuracy gap for candidate sentences that include these KG paths.

LogBERT slightly underperforms KENLI+E for CMV, but it outperforms KENLI+E for Kialo, possibly because LogBERT is trained on arguments from Kialo and may learn useful linguistic information in Kialo. While KENLI+E has relatively high recall, LogBERT is notable for high precision compared to the other models. This is somewhat counterintuitive because LogBERT uses four logical relations between two statements, which one might expect to improve recall by capturing a broad range of mechanisms for contradiction. In reality, however, LogBERT seems to make conservative predictions based on whether strong signals exist for the logical relations.
Combining the two models may result in high recall by incorporating different KGs and, at the same time, improve precision by incorporating different types of signals (e.g., sentiment). For example, KENLI could be pretrained on the four logical mechanisms in the same way that LogBERT is. Alternatively, we could incorporate the KE Net in the middle of LogBERT. We leave this direction to future work.

We conducted a further analysis on how differently KENLI+E and LogBERT behave. We find that LogBERT excels when a pair of statement and candidate sentence has strong signals for four logical relations---textual contradiction, negative sentiment, obstructive causal relation, and refuting normative relation---that have been found to have high correlations with LogBERT's decision of `contradiction' \cite{Jo:tacl21}. For instance, for CMV, when we focus only on the pairs whose candidate sentences express negative sentiment toward their target statements with probability greater than 0.5, LogBERT's recall and F1-score substantially increase to 96.5 and 65.3, respectively, whereas KENLI+E's recall and F1-score drop to 63.1 and 56.9. This means that negative sentiment toward the target statement is a useful cue for counterevidence, and LogBERT uses this signal better than KENLI. The caveat, however, is LogBERT's F1-score drops significantly to 49.9 for candidate sentences not expressing negative sentiment, while KENLI+E's F1-score remains stable at 58.4. A similar pattern occurs for obstructive causal relation and textual contradiction, and for Kialo as well. More details are in Appendix~\ref{app:kenli_vs_logbert}.

% In terms of other NLI models that incorporate KGs, we want to emphasize that our method substantially outperforms previous models, which supports the novelty and strength of our model design. We referred to three such papers recently published (Chen et al. 2018, Song et al., 2020, Wang et al., 2020) and their reported accuracy scores for MNLI / MNLI-MM are as follows: [Chen] 77.2 / 76.4, [Song] 76.1 / 74.2, [Wang] 79.1 / 78.2, [KENLI] 87.3 / 87.2. We will add this comparison and full references in the final version.

According to Kendall's $\tau$, LogBERT shows the best alignment with human judgments on the validity scores of candidate sentences among the four models. However, correlations overall are rather weak, ranging between 0.002 and 0.045. 
One possible reason is that the models are trained for binary classification, not to predict the degree of validity. As a result, 75\% of KENLI+E's probability scores are skewed toward $>0.9$ or $<0.1$ versus only 10\% of human scores (after normalized between 0 and 1). Therefore, KENLI+E's probability scores seem to capture something different than human scores by design. KENLI may have a better alignment with human scores if it is trained explicitly on the degree of validity using regression, though collecting such data would be expensive. Another reason for the low correlation is that sometimes a candidate sentence's validity is vague due to limited context (e.g., Statement: \textit{However, the similarities end there.} / Candidate Sentence: \textit{However, the similarities do not end there.}). In such cases, human scores tend to lie in the middle area (0.33 for this example), whereas KENLI still makes a confident decision (0.99 for this example) within the limited context. We also find that KENLI tends to overpredict `valid' as a candidate sentence and the target statement share more words. Such an overreliance on overlapping words could exacerbate the misalignment with human judgments.

%The reason that the models have higher accuracy for Kialo than for CMV is not clear. We analyzed accuracy by the length of refuted statements and by whether the refuted statement is normative or not, but we did not find conclusive evidence that they are important factors for accuracy. 

We conducted further analyses on the utility of document types. Previous work on counterargument generation and fact extraction/verification relies heavily on Wikipedia and sometimes news articles. However, we find that valid counterevidence resides in more various sources. While knowledge archives (e.g., Wikipedia, lectures) take the highest proportion (27--37\%), counterevidence resides in mainstream news, personal blogs, research journals, etc. as well (Figure~\ref{fig:evidence_domain_dist} in Appendix~\ref{app:utility_document_types}). Moreover, if we break down model accuracy into different document types, the models are more reliable (i.e., achieve higher F1-scores) for specialized magazines and Q\&A forums than knowledge archives (Figure~\ref{fig:evidence_domain_accs} in Appendix~\ref{app:utility_document_types}). If we break down model accuracy by search methods, Wikipedia achieves lower scores than Bing and Google across all metrics (Table~\ref{tab:evidence_accs_search} in Appendix~\ref{app:utility_search_methods}). These results suggest that counterargument generation and fact extraction/verification should consider more diverse sources of evidence beyond Wikipedia. See Appendices~\ref{app:utility_document_types} and \ref{app:utility_search_methods} for more details.

In \S{\ref{sec:intro}}, we assumed the scenario where one makes a counterargument by first detecting attackable points in the target argument and then retrieving counterevidence to those points. To see the feasibility of automating this pipeline, we took an existing model that aims to detect attackable sentences in arguments \cite{Jo2020:attackable} and analyzed whether this model can identify sentences that have counterevidence according to our collected data. We find that attackability scores predicted by this model tend to be higher for argument sentences for which we were able to find counterevidence (Figure~\ref{fig:attack} in Appendix~\ref{app:retrieval_results}). This result suggests that computational models for attackability detection and our counterevidence retrieval system could create synergy to fully automate counterargument generation. See Appendix~\ref{app:attackability} for more details.

%% file: texes/5_Conclusion.tex
In this paper, we built a counterevidence retrieval system. To retrieve counterevidence that involves complex inference, we presented a knowledge-enhanced NLI model with specific focus on causality- and example-based inference. The NLI model demonstrates improved performance for NLI tasks, especially for instances that require the targeted inference. Integrating the NLI model into the retrieval system further improves counterevidence retrieval performance, especially recall, showing the effectiveness and utility of our method of incorporating knowledge graphs in NLI.

%% file: texes/Appendix.tex
\section*{Appendix}

\section{Annotation Tasks}

\subsection {Annotation Principle}

For all annotation tasks, we recruited annotators on Amazon Mechanical Turk (MTurk).
Participants should meet the following qualifications: (1) residents of the U.S., (2) at least 500 HITs approved, and (3) HIT approval rate greater than 97\%.
Each HIT includes several questions and one attention question.
The attention question asks the annotator to select a specific option, and we rejected and discarded a HIT if the annotator failed the attention question. For all annotation tasks, the annotation manuals are publicly available.

% 각 task 별 annotator 수
% <valid>
% cmv: 3677
% kialo: 2711
% domain: 7900
% example: 21980

% <invalid>
% cmv: 166
% kialo: 95
% domain: 2081
% example: 7740

\subsection{Annotation of Example-Based NLI data\label{sec:annot_example}}
This section describes our method for synthetic building of example-based NLI data that was augmented with the public NLI datasets in our experiments.
The entire process consists of two steps. First, we generate synthetic example-based pairs using a pretrained language model (\S{\ref{sec:app_how_example_based_nli}}). Next, we annotate their labels using MTurk (\S{\ref{sec:app_example_based_human_annotation}}).

\subsubsection{Generating Synthetic NLI Pairs}
\label{sec:app_how_example_based_nli}
We synthetically generate example-based NLI pairs as follows.
Given a pair of $P$ and $H$ in the public datasets in Table~\ref{tab:datasets}, we modify $P$ to $P'$ by replacing an entity that occurs in both $P$ and $H$ with an incoming entity on WikidataEx.
For example, in the following pair
\qargutwo{a breakdancer man is performing for the kids at school}{a man is break dancing at a school}
``school'' occurs in both $P$ and $H$, so we may generate $P'$ by replacing ``school'' with an instance of the school (e.g., ``preschool'') based on WikidataEx. 
To avoid broken or implausible sentences, we retain $P'$ only if its perplexity is lower than or equal to that of $P$ based on GPT2. 
Table \ref{tab:example_based_statements_examples} shows examples of synthetically generated $P'$ and their perplexity.
$P$ is the original statement from the SNLI dataset, and $P_1'$--$P_4'$ are generated statements after the entity ``school'' is replaced.
The perplexity of $P_1'$ and $P_2'$ is lower than that of the original statement $P$, so we pair each of them with $H$ and add the pairs to our synthetic NLI data. 
However, $P_3'$ and $P_4'$ are discarded because their perplexity is higher than that of $P$.
%We create 3.6 billion $P'$ sentences by replacing the entity and select 1.6 billion statements that are lower than or equal to that of the original statement $P$ to make new $(P', H)$ pairs.

% 3,620,211,370
% Through this method, we select 45.7\% of all generated example-based statements.
% 이렇게 만들고 나서 제가 유하늘씨에게 데이터 전체를 전달한 것으로 기억합니다만 아마 개수의 문제로 전부 다 tagging은 하지 않은 것 같아요.
% 혹시 몇 개를 하셨는지 알려주실 수 있으실까요?
% 6031개입니다 (train 702개, val 702개, test 4627개)

\begin{table}[t]
    \centering
    \small
    \begin{tabularx}{\linewidth}{lXr}
        \toprule
        ID & Statement & Perplexity \\
        \midrule
        $P$  & a breakdancer man is performing for the kids at school & 3.08 \\
        $P_1'$ & a breakdancer man is performing for the kids at licensed victuallers' school & 2.67 \\
        $P_2'$ & a breakdancer man is performing for the kids at preschool
         & 2.95 \\
        $P_3'$ & a breakdancer man is performing for the kids at boys republic & 3.09 \\
        $P_4'$ & a breakdancer man is performing for the kids at language teaching & 3.10 \\
        \bottomrule
    \end{tabularx}
    \caption{Examples of generated example-based statement and its perplexity measured by GPT2.}
    \label{tab:example_based_statements_examples}
\end{table}

\subsubsection{Label Annotation}
\label{sec:app_example_based_human_annotation}
\begin{figure*}[!t]
    \centering
    \includegraphics[width=\linewidth]{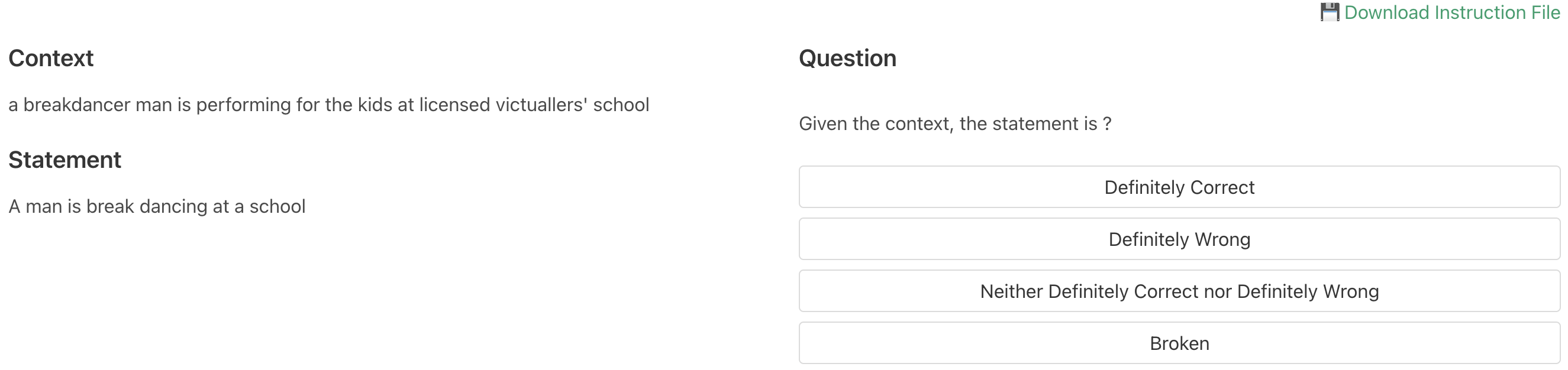}
    \caption{Example-based NLI labeling page.}
    \label{fig:eval_example}
\end{figure*}

For each of the generated NLI pairs, we ask annotators whether $H$ is correct or wrong given the context $P'$. They can choose from the four options: definitely correct (entail), definitely wrong (contradict), neither definitely correct nor definitely wrong (neutral), and broken English (Figure \ref{fig:eval_example}).
Each HIT consists of 10 pairs and one attention question. Each pair is labeled by three annotators and is discarded if the three annotators all choose different labels.

\subsubsection{Analysis}
\begin{table}[t]
    \centering
    \small
    \begin{tabularx}{\linewidth}{lcCCC}
        \toprule
        & & \multicolumn{3}{c}{Original Label} \\
        \cmidrule{3-5}
        & & Entail & Neutral & Contradict \\
        \midrule
        \multirow{5}{*}{\rotatebox{90}{New Label}}
            & Entail & 1,698 & 548 & 373\\
        & Neutral & 228 & 543 & 139\\
        & Contradict & 151 & 302 & 1,030\\
        & Broken & 20 & 15 & 24\\
        & No Majority & 336 & 333 & 291\\
        \bottomrule
    \end{tabularx}
\caption{Confusion matrix of example-based NLI data labels.}
\label{tab:example_result}
\end{table}

To see how the labels of the generated pairs $(P',H)$ differ from the labels of their original pairs $(P,H)$, we manually analyzed 6,031 pairs (Table~\ref{tab:example_result}). 
Only 59 sentences were labeled as broken, meaning that our GPT2-based generation method effectively generates sensible statements $P'$. 
Most original pairs of entailment and contradiction keep their labels, but many of originally neutral pairs turn to either entailment or contradiction after entity replacement.

% %\subsection{Results\label{sec:retrieval_results}}

\subsection{Annotation of Evidence Validity\label{sec:annot_evidence}}
\begin{figure*}[!t]
    \centering
    \includegraphics[width=\linewidth]{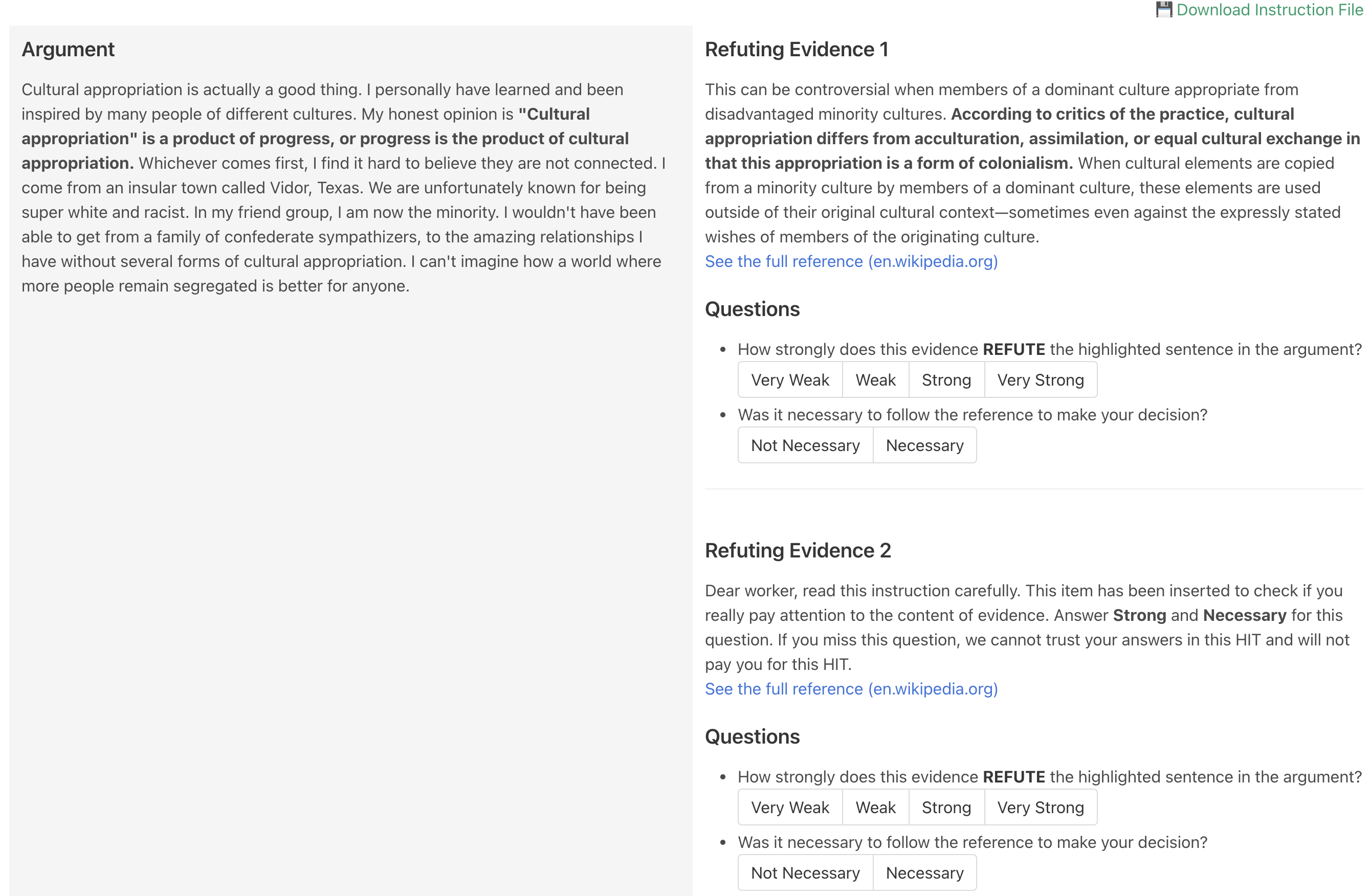}
    \caption{CMV evaluation page. Refuting Evidence 2 is an attention question.}
    \label{fig:eval_cmv}
\end{figure*}

\begin{figure*}[!t]
    \centering
    \includegraphics[width=\linewidth]{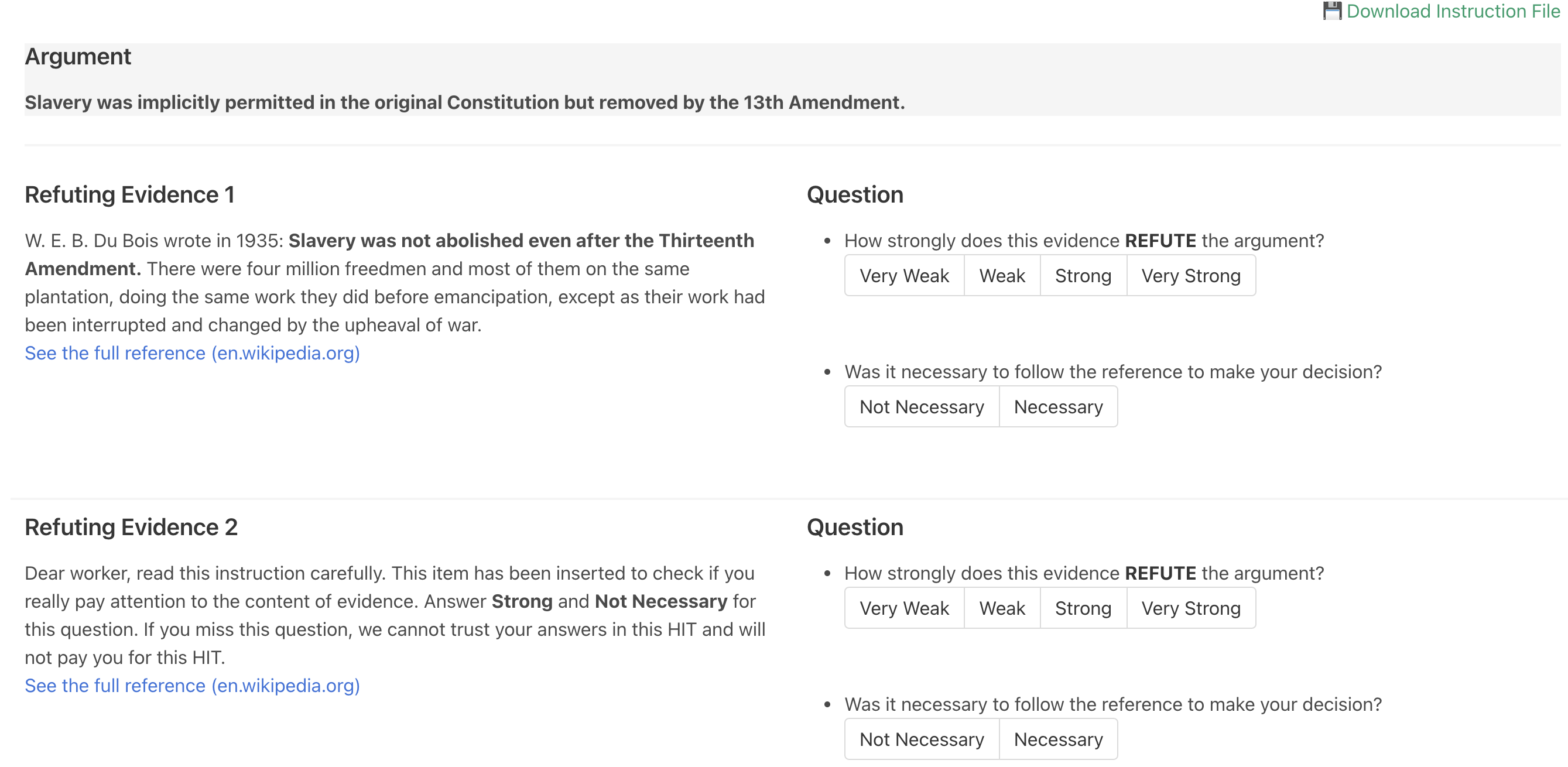}
    \caption{Kialo evaluation page. Refuting Evidence 2 is an attention question.}
    \label{fig:eval_kialo}
\end{figure*}

In this task, annotators were asked to mark how strongly each counterevidence candidate sentence refutes the statement it attempts to refute (i.e., statements from CMV or Kialo). The four options of strength are `very weak', `weak', `strong', and `very strong', with corresponding scores 0, 1, 2, and 3 (Figures~\ref{fig:eval_cmv} and \ref{fig:eval_kialo}). For each statement from CMV, the entire post is displayed with the target statement highlighted so the annotator can consider the context of the statement when making a decision. For each candidate sentence, the annotators should also answer whether reading the source document is necessary to make a judgment.

Each HIT includes four statements to refute, along with at most four candidate counterevidence sentences for each statement, and one attention question.
Each candidate sentence was labeled by two annotators.
If a candidate sentence was labeled as both `very weak' and `very strong', we treated the labels as unreliable (146 candidates in 131 sentences from CMV, 71 candidates in 65 statements from Kialo) and allocated a third annotator.
We average their scores, which becomes the candidate sentence's final strength.
The average variance of scores for each candidate sentence is 0.48, meaning that annotators on average have a score difference less than 1 point.
%Evaluation on both CMV and Kialo shows Cohen's Kappa with 0.22 which indicates fair agreement between annotators.

\subsection{Annotation of Document Types\label{sec:annot_cand_types}}

\begin{figure*}[!t]
    \centering
    \includegraphics[width=\linewidth]{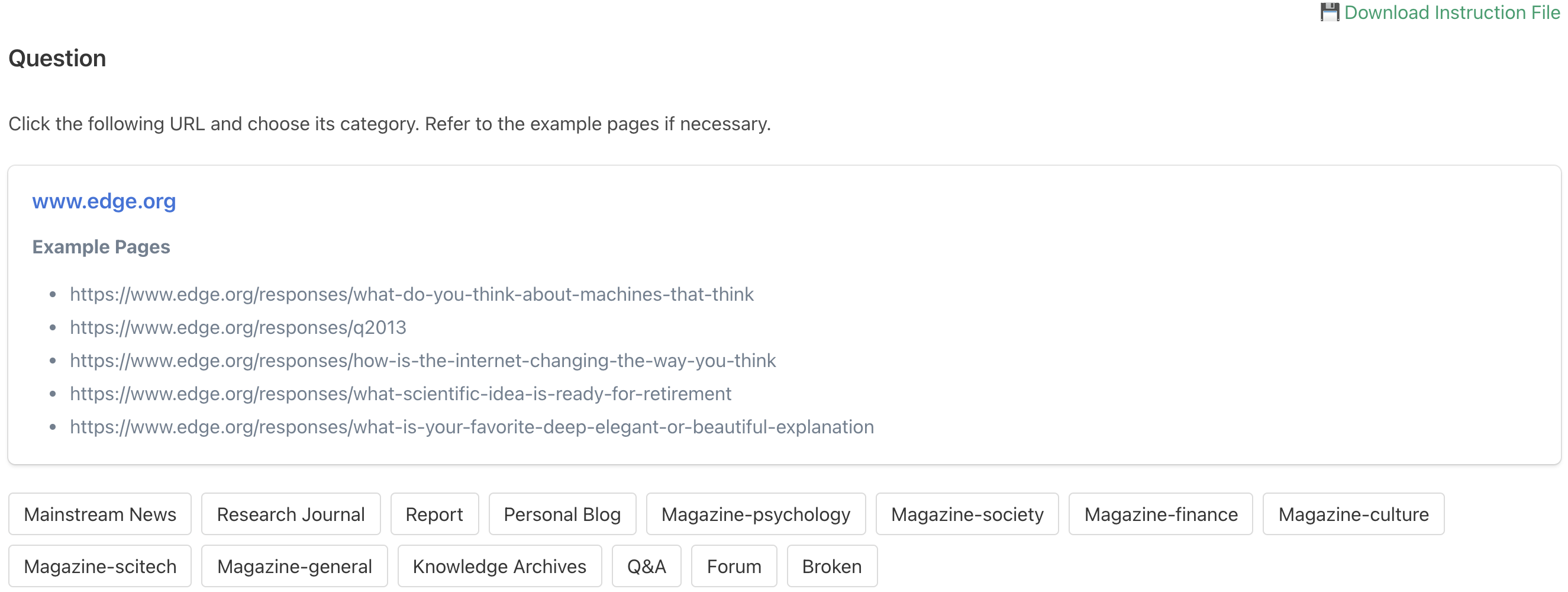}
    \caption{Document type annotation page.}
    \label{fig:eval_domain}
\end{figure*}

\begin{table*}[!t]
    \centering
    \small
    \begin{tabularx}{\linewidth}{lp{7cm}X} 
        \toprule
        Type & Description & Examples \\
        \midrule
        Mainstream News & Mainstream news about daily issues and general topics. & www.cnn.com, www.bbc.com \\
        Research Journal & Peer-reviewed papers or dissertations. & link.springer.com, www.nature.com \\
        Report & Surveys, statistics, and reports. Should be a source of substantial data rather than a summary of reports. & www.whitehouse.gov, www.irs.gov, www.cdc.gov \\
        Personal Blog & Personal blogs. & medium.com, jamesclear.com \\
        \midrule
        Magazine--Psychology & Magazines about psychology, mental health, relationships, family. & www.psychologytoday.com \\
        Magazine--Society & Magazines about social and political issues. & www.hrw.org, www.pewresearch.org \\
        Magazine--Finance & Magazines about finance, business, management. & www.hbr.org \\
        Magazine--Culture & Magazines about culture, education, entertainment, fashion, art. & www.vulture.com \\
        Magazine--Scitech & Magazines about science, medicine, technology. & www.techdirt.com, www.webmd.com \\
        Magazine--General & Magazines about multiple domains. &  thedickinsonian.com \\
        \midrule
        Knowledge Archives & Information archives for knowledge transfer, such as encyclopedias, books, dictionaries, lectures. & plato.stanford.edu, quizlet.com, www.wikihow.com \\
        Q\&A & Question and answering platforms. & stackoverflow.com, www.quora.com\\
        Forum & Forums for opinion sharing and reviews. & www.reddit.com, www.debate.org \\
        \midrule
        Broken & URLs are not accessible. & \\
        \bottomrule
    \end{tabularx}
    \caption{Evidence document types.}
    \label{tab:evidence_domains}
\end{table*}

In this task, we annotate the type of source document for each candidate sentence. Each annotator was shown the network location identifier of a URL (e.g., ``www.cnn.com'', ``docs.python.org'') and asked to choose the type of the site from 14 categories (Table~\ref{tab:evidence_domains} and Figure~\ref{fig:eval_domain}).
Total 1,987 unique location identifiers were annotated. Each HIT consists of 10 identifiers and one attention question.
Each identifier was annotated until two annotators chose the same category. If there was no such category for five annotators, we selected the decision of the most ``trustworthy'' annotator, who had the highest rate of decisions selected for other identifiers.

\subsection{Ethical Considerations on Human Annotation}
\begin{table*}[th]
    \small
    \centering
    \begin{tabularx}{\linewidth}{lCCCC}
        \toprule
          Task & \# Questions/HIT & Time/HIT (secs) & Wage/HIT & Expected Hourly Wage \\
          \midrule
        Example-based NLI & 10 & 324\hide{5'24"} & \$0.7 & \$7.78 \\
        Evidence Validity -- CMV & 4 & 247\hide{4'07"} & \$0.5 & \$7.28  \\
        Evidence Validity -- Kialo & 4 & 240\hide{4'00"} & \$0.5 & \$7.50  \\
        Document Type & 10 & 351\hide{3'51"} & \$0.5 & \$7.79  \\
        \bottomrule
    \end{tabularx}
    \caption{Expected hourly wage of each annotation task. All wages are over the federal minimum wage in the U.S. (\$7.25). The number of questions per HIT does not include attention questions.}
    \label{tab:expected_hourly_wage}
\end{table*}

% Ethics Review Questions
% \url{https://2021.emnlp.org/call-for-papers/ethics-review-questions}

% Emily M. Bender's question:
% How did you ensure the crowdworkers were paid a fair wage for their labor?
% \url{https://twitter.com/emilymbender/status/1288214202801217536}

% Related docs: \url{https://medium.com/ai2-blog/crowdsourcing-pricing-ethics-and-best-practices-8487fd5c9872} \& \cite{whiting2019fair}

% Factors: Pricing, Privacy, and Transparency.

We consider ethical issues on our annotation tasks.
The first consideration is fair wages.
We compute the average time per HIT based on a small pilot study, and set the wage per HIT to be above the federal minimum wage in the U.S. (\$7.25\footnote{\url{https://www.dol.gov/general/topic/wages/minimumwage}}).
Table~\ref{tab:expected_hourly_wage} shows that the expected hourly wage is higher than the federal minimum wage for all the annotation tasks.

We also preserve the privacy of crowdworkers.
We do not ask for their personal information, such as names and gender.
We collect Worker IDs to map each HIT result with the annotator and to accept or reject their work on MTurk.
But the Worker IDs are discarded afterward to preserve their privacy.

Our annotation tasks are upfront and transparent with annotators.
We provide the instruction manual of each task at the starting page, which informs the annotators of various task information, such as an estimated time needed for the task.
Some annotators complained when their work was rejected.
We generally responded within a business day with evidence of our decision (i.e., their failure at the attention question).

% \citet{whiting2019fair} suggest that \$15/hr minimum wage was the fairest option.

\section{Example NLI Pairs\label{app:nli_examples}}
Table~\ref{tab:nli_examples} shows example NLI pairs that are classified correctly by only one model.

\begin{table*}[p]
    \small
    \centering
    \begin{tabularx}{\linewidth}{lX}
        \toprule
        Correct Model & NLI Pair \\
        \midrule
        \multirow{3}{*}{AdaptBERT+C} & \textbf{\textit{P}:} How to blow a bubble with bubblegum<br>Buy some bubblegum. You can buy gum at pretty much every corner store. Chewing gums can be used to make bubbles, but they won't be as big, and they'll usually pop too easily. / \textbf{\textit{H}:} Kids can learn to blow bubbles. \\
         & \textbf{\textit{P}:} Letters in support or condemnation of the QES program (though one may assume they will insist on programme ) should be addressed to Mrs Anne Shelley, Secretary, Queen's English Society, 3 Manor Crescent, Guildford GU2 6NF, England. / \textbf{\textit{H}:} Mrs. Anne Shelley is in charge of the QES program. \\
         & \textbf{\textit{P}:} A woman in black walks down the street in front of a graffited wall. / \textbf{\textit{H}:} A young woman is standing and staring at a painted mural. \\
        \midrule
        \multirow{3}{*}{AdaptBERT+E} & \textbf{\textit{P}:} Donna Noble is a fictional character in the long-running British science fiction television series Doctor Who . Portrayed by British actress and comedian Catherine Tate , she is a companion of the Tenth Doctor ( David Tennant ) . / \textbf{\textit{H}:} Donna Noble is the therapist of the Doctor. \\
         & \textbf{\textit{P}:} Blond boy in striped shirt on the swing set. / \textbf{\textit{H}:} A boy in a blue striped shirt is on the swing set. \\
         & \textbf{\textit{P}:} The diocese of Vannida (in Latin: Dioecesis Vannidensis) is a suppressed and titular See of the Roman Catholic Church. It was centered on the ancient Roman Town of Vannida, in what is today Algeria, is an ancient episcopal seat of the Roman province of Mauritania Cesariense. / \textbf{\textit{H}:} The diocese of Vannida is located in Europe \\
        \midrule
        \multirow{3}{*}{K-BERT+C} & \textbf{\textit{P}:} Second, a clue may name a class of objects which includes the answer, like bird for COCK. / \textbf{\textit{H}:} Clues may not be categorical. \\
         & \textbf{\textit{P}:} Asia has the highest number of child workers, but Sub-Saharan Africa has the highest proportion of working children relative to population. / \textbf{\textit{H}:} There are more children in Asia than there are in Sub-Saharan Africa. \\
         & \textbf{\textit{P}:} Duende meant `hobgoblin,' `sprite,' or `ghost' in Spanish for a long time, but it is not known when it acquired its artistic coloration. / \textbf{\textit{H}:} Duende has had many meanings, but all were similar. \\
        \midrule
        \multirow{3}{*}{K-BERT+E} & \textbf{\textit{P}:} Well, the issue before this Court, I hasten to say, as I said before, is only whether, once the Congress makes that judgment, it can ever change it retrospectively. The issue before this Court is not whether, in the future, a certain length of time would be appropriate. / \textbf{\textit{H}:} Congress will change the decision of a judgement without considering when the decision was made. \\
         & \textbf{\textit{P}:} John Goodman . His other film performances include lead roles in The Babe ( 1992 ) , The Flintstones ( 1994 ) and 10 Cloverfield Lane ( 2016 ) and supporting roles in Coyote Ugly ( 2000 ) , The Artist ( 2011 ) , Extremely Loud and Incredibly Close ( 2011 ) , Argo ( 2012 ) , Flight ( 2012 ) , The Hangover Part III ( 2013 ) , and Patriots Day ( 2016 ) . / \textbf{\textit{H}:} John Goodman played Babe Ruth in The Babe. \\
         & \textbf{\textit{P}:} Man chopping wood with an axe. / \textbf{\textit{H}:} The man is outside. \\
        \midrule 
        \multirow{3}{*}{RoBERTa} & \textbf{\textit{P}:} Two men wearing dirty clothing are sitting on a sidewalk with their dog and begging for money using a cardboard sign. / \textbf{\textit{H}:} These two illiterate men convinced their dog to write a panhandling sign for them. \\
         & \textbf{\textit{P}:} Sure enough, there was the chest, a fine old piece, all studded with brass nails, and full to overflowing with every imaginable type of garment.  / \textbf{\textit{H}:} The chest wasn't big enough to completely contain all of the garments. \\
         & \textbf{\textit{P}:} A man is spinning a little girl in the air above his head. / \textbf{\textit{H}:} A man is carrying a little girl off the ground. \\
        \midrule
        \multirow{3}{*}{KENLI+C} & \textbf{\textit{P}:} "Nine Lives" is a song by American hard rock band Aerosmith. It was released in 1997 as the lead single and title track from the album "Nine Lives". The song was written by lead singer Steven Tyler, guitarist Joe Perry, and songwriter Marti Frederiksen. The song is four minutes, one second long. All the high-caliber guitar solos are played by Brad Whitford. / \textbf{\textit{H}:} The song "Nine Lives" was released in 1997 and performed only by Brad Whitford. \\
         & \textbf{\textit{P}:} The last stages of uploading are like a mental dry-heave. / \textbf{\textit{H}:} Uploading your consciousness feels like a mental dry-heave in the final stages. \\
         & \textbf{\textit{P}:} Cheerleaders in blue performing on a football field underneath a yellow football goal post. / \textbf{\textit{H}:} A group of girls shake pom poms. \\
        \midrule
        \multirow{3}{*}{KENLI+E} & \textbf{\textit{P}:} The Ron Clark Story is a 2006 television film starring Matthew Perry. The film is based on the real-life educator Ron Clark. It follows the inspiring tale of an idealistic teacher who leaves his small hometown to teach in a New York City public school, where he faces trouble with the students. The film was directed by Randa Haines, and was released directly on television. / \textbf{\textit{H}:} Ron Clark's hometown was not New York City. \\
         & \textbf{\textit{P}:} These social encounters oer children many opportunities to hear people refer to their own mental states and those of others and, therefore, to observe dierent points of view. / \textbf{\textit{H}:} The social encounters give kids a lot of chances to hear people talk about their mental states and how things make them feel. \\
         & \textbf{\textit{P}:} An astute mother I observed in the grocery store had her 3-year-old son, Ricky, reach for items on the shelf and put them in the cart. / \textbf{\textit{H}:} Ricky was a good boy and followed his mother doing nothing else.  \\
        \bottomrule
    \end{tabularx}
    \caption{NLI pairs that are correctly classified by only one model. `Correct Model' is the only model that correctly classifies the NLI pairs to the right.}
    \label{tab:nli_examples}
\end{table*}

\section{Counterevidence Retrieval System\label{sec:deseption}}
\paragraph{Document Retrieval:} 
In this stage, documents that may contain counterevidence are retrieved. Given a statement to verify, DeSePtion retrieves candidate documents from Wikipedia in four ways: (1) using named entities in the statement as queries for the wikipedia library\footnote{\url{https://pypi.org/project/wikipedia/}}, (2) using the statement as a query for Google, (3) TF-IDF search using DrQA \cite{Chen:2017drqa}, and (4) some heuristics. Note that all documents are from Wikipedia, in accordance with the FEVER task.

We make several adaptations that better suit our task. First, in addition to Wikipedia articles, we also retrieve web documents using Microsoft Bing and Google (wikipedia pages are excluded from their search results). The three sources provide documents with somewhat different characteristics, and we compare their utility in Appendix~\ref{app:retrieval_results}. Second, we use the Spacy Entity Linker to retrieve the articles of Wikidata entities linked to the statement. And for each linked entity, we additionally sample at most five of their instance entities and the corresponding articles. These expanded articles potentially include counterexamples to the statement\footnote{We considered retrieving web documents in a similar way, using query expansion, but ended up not doing it. One reason is that search engines already include example-related documents to some extent. For instance, for the query ``Vegan diets can cause cancer'', Bing returns a document with the title ``Can the Keto and Paleo Diets Cause Breast Cancer?''. Another practical reason is that query expansion requires arbitrarily many search transactions that are beyond the capacity of our resources.}. Lastly, we do not use the heuristics.

\paragraph{Document Ranking:}
Given a set of candidate documents, DeSePtion ranks them using a pointer net combined with fine-tuned BERT. First, BERT is trained to predict whether each document is relevant or not, using the FEVER dataset; it takes the concatenation of the page title and the statement to verify as input. The output is used as the embedding of the document. Next, a pointer net takes these embeddings of all documents and sequentially outputs pointers to relevant documents.

In our adaptation, we use RoBERTa in place of BERT. More importantly, we use search snippets in place of page titles to take advantage of the relevant content in each document provided by search engines. The ranker is still trained on the FEVER dataset, but since it does not include search snippets, we heuristically generate snippets by concatenating the title of each Wikipedia page with its sentence that is most similar to the statement\footnote{We combine all token embeddings in the last layer of RoBERTa and measure the cosine similarity between these vectors.}. This technique substantially improves document relevance prediction on the FEVER dataset by 7.4\% F1-score points. For web documents, we use search snippets provided by Bing and Google.

The number of retrieved documents varies a lot depending on the search method; the Google API retrieves much fewer documents than Wikipedia and Bing in general. Since this imbalance makes it difficult to compare the utility of the different search methods, we make the number of candidate documents the same across the methods, by ranking documents from different search methods separately and then pruning low-ranked documents of Wikipedia and Bing. This process lets the three methods have the same average number of candidate documents per statement ($\sim$8).

\paragraph{Sentence Selection:}
DeSePtion considers all sentences of the ranked documents. However, web documents are substantially longer than Wikipedia articles in general, so it is computationally too expensive and introduces a lot of noise to process all sentences. Therefore, for each statement to verify, we reduce the number of candidate sentences by selecting the top 200 sentences (among all ranked documents) whose RoBERTa embeddings have the highest cosine similarity to the statement.

\paragraph{Relation Prediction:}
In this stage, we classify whether each candidate sentence is valid counterevidence to the statement to verify. Here, instead of DeSePtion, we simply use an NLI model as-is. The reason is that the main goal of DeSePtion is to predict the veracity of a statement, rather than whether each sentence supports or refutes the statement. Thus, it assumes that once a statement is found to be supported or refuted, considering more sentences results in the same prediction. This assumption is justified for the FEVER task, where a statement cannot be both supported and refuted. In real-world arguments, however, a statement can be both supported and refuted, and our goal is to find refuting sentences. We compute the probability score that each sentence contradicts the statement and rank the sentences by these scores. This simple approach has been found to be effective in information retrieval~\cite{Dai:2019.bert}.

\section{Comparison between KENLI+E and LogBERT\label{app:kenli_vs_logbert}}
In order to better understand how differently KENLI+E and LogBERT behave, we broke down the performance of the models based on how strongly a pair of statement and candidate sentence signals the four logical relations---textual contradiction, negative sentiment, obstructive causal relation, and refuting normative relation---that have high correlations with LogBERT's decision of `contradiction'~\cite{Jo:tacl21}. More specifically, when LogBERT takes a pair of statement and candidate sentence, it can compute the probabilities of these four logical relations using its pretrained classification layers. For each of these relations, we split input pairs into two groups: pairs with probability greater than 0.5 and the other. Our hypothesis is that LogBERT performs well on the pairs in the first group, because they have a strong signal associated with `contradiction' that LogBERT learns during pretraining.

Table~\ref{tab:accs_breakdown} shows the breakdown of model accuracy. As expected, LogBERT usually achieves the highest recall and F1-score among all the models for input pairs that have strong signals of the logical mechanisms ($P > 0.5$). This pattern is pronounced for negative sentiment and obstructive causal relation. In contrast, LogBERT's F1-score drops substantially when such signals are missing ($P \leq 0.5$), whereas KENLI+E's performance is more stable regardless of those signals. This result implies the overreliance of LogBERT on the four logical relations, which is helpful when such relations exist in input pairs but rather harmful otherwise. 

\begin{table*}[t]
    \centering
    \small
    \begin{tabularx}{\linewidth}{ll CCC p{0.5mm} CCC p{2mm} CCC p{0.5mm} CCC}
        \toprule
         & & \multicolumn{7}{c}{CMV} & & \multicolumn{7}{c}{Kialo} \\
        \cmidrule{3-9} \cmidrule{11-17}
         & & \multicolumn{3}{c}{$P > 0.5$} & & \multicolumn{3}{c}{$P \leq 0.5$} & & \multicolumn{3}{c}{$P > 0.5$} & & \multicolumn{3}{c}{$P \leq 0.5$} \\
        \cmidrule{3-5} \cmidrule{7-9} \cmidrule{11-13} \cmidrule{15-17}
         & & Prec & Recl & F1 & & Prec & Recl & F1 & & Prec & Recl & F1 & & Prec & Recl & F1 \\
        \midrule
        \rotate{4}{\shortstack{Textual\\Contradict}} & RoBERTa & 50.2 & 60.9 & 55.0 &    & 45.6 & 68.6 & 54.8 &   & 60.4 & 53.8 & 56.9 &    & 53.9 & 64.0 & 58.5 \\
         & KENLI+C & 51.2 & 65.8 & 57.6 &    & 44.7 & 63.7 & 52.5 &   & \textbf{61.6} & 59.2 & 60.3 &    & \textbf{54.7} & 69.0 & 61.1 \\
         & KENLI+E & \textbf{51.5} & 69.7 & 59.2 &    & 44.9 & \textbf{74.3} & \textbf{56.0} &   & 60.8 & 62.1 & 61.4 &    & 53.2 & \textbf{72.0} & \textbf{61.2} \\
         & LogBERT & \textbf{51.5} & \textbf{83.4} & \textbf{63.7} &    & \textbf{50.7} & 21.4 & 30.1 &   & 60.8 & \textbf{86.8} & \textbf{71.5} &    & 53.0 & 20.2 & 29.2 \\
        \midrule
        \rotate{4}{\shortstack{Negative\\Sentiment}} & RoBERTa & 51.3 & 57.0 & 54.0 &    & 47.3 & 66.4 & 55.3 &   & 61.6 & 44.6 & 51.7 &    & 56.4 & 65.7 & 60.7 \\
         & KENLI+C & 51.0 & 60.3 & 55.3 &    & 48.0 & 67.1 & 56.0 &   & \textbf{62.8} & 49.7 & 55.5 &    & 57.4 & 71.0 & 63.4 \\
         & KENLI+E & \textbf{51.8} & 63.1 & 56.9 &    & 47.8 & \textbf{74.9} & \textbf{58.4} &   & 61.5 & 52.3 & 56.5 &    & 56.4 & \textbf{74.2} & \textbf{64.1} \\
         & LogBERT & 49.4 & \textbf{96.5} & \textbf{65.3} &    & \textbf{53.3} & 46.9 & 49.9 &   & 60.8 & \textbf{96.9} & \textbf{74.7} &    & \textbf{58.8} & 44.7 & 50.8 \\
        \midrule
        \rotate{4}{\shortstack{Obstructive\\Causal}} & RoBERTa & 52.5 & 60.5 & 56.2 &    & 47.8 & 64.0 & 54.7 &   & 61.5 & 53.6 & 57.3 &    & 57.3 & 57.7 & 57.5 \\
         & KENLI+C & 52.5 & 66.9 & 58.8 &    & 48.3 & 64.8 & 55.4 &   & \textbf{61.9} & 59.8 & 60.9 &    & 58.5 & 62.7 & 60.5 \\
         & KENLI+E & \textbf{54.5} & 69.8 & 61.2 &    & 48.2 & \textbf{71.5} & \textbf{57.6} &   & \textbf{61.9} & 64.0 & 63.0 &    & 57.2 & \textbf{65.4} & \textbf{61.0} \\
         & LogBERT & 52.6 & \textbf{100.0} & \textbf{68.9} &    & \textbf{51.1} & 56.5 & 53.7 &   & 58.4 & \textbf{100.0} & \textbf{73.8} &    & \textbf{60.5} & 59.4 & 59.9 \\
        \midrule
        \rotate{4}{\shortstack{Refuting\\Normative}} & RoBERTa & 46.6 & 62.9 & 53.5 &    & \textbf{49.6} & 64.0 & 55.9 &   & 59.3 & 61.6 & 60.4 &    & 56.9 & 53.8 & 55.3 \\
         & KENLI+C & 47.8 & 64.1 & 54.8 &    & 49.5 & 65.7 & 56.5 &   & 60.8 & 66.3 & 63.4 &    & 57.8 & 59.4 & 58.6 \\
         & KENLI+E & 48.5 & 70.5 & 57.4 &    & 49.1 & \textbf{71.9} & \textbf{58.4} &   & 59.0 & \textbf{69.6} & 63.9 &    & 57.2 & 62.1 & 59.6 \\
         & LogBERT & \textbf{53.7} & \textbf{71.0} & \textbf{61.1} &    & 49.5 & 55.4 & 52.3 &   & \textbf{61.7} & 67.0 & \textbf{64.3} &    & \textbf{58.8} & \textbf{65.7} & \textbf{62.1} \\
        \bottomrule
    \end{tabularx}
    \caption{Breakdown of model accuracy by the strength of logical relations in input pairs. The first column indicates a logical mechanism that is associated with LogBERT's decision of `contradiction'. `$P > 0.5$ ($\leq 0.5$)' indicates input pairs whose probability of the logical mechanism is greater than (less than or equal to) 0.5. For each mechanism, bold numbers indicate the highest score for each metric.}
    \label{tab:accs_breakdown}
\end{table*}

\section{In-Depth Analyses of Evidence Retrieval\label{app:retrieval_results}}

\subsection{Utility of Search Methods\label{app:utility_search_methods}}
One difference between our system and prior work is that we retrieved web documents using Bing and Google, whereas no prior work did that to our knowledge. Hence, comparing candidate sentences from Wikipedia, Bing, and Google will shed light on the usefulness of the search engines and inform future system designs. Table~\ref{tab:evidence_accs_search} shows candidate sentences retrieved from Bing and Google generally achieve higher F1-scores than those from Wikipedia. While Wikipedia provides comparably good recall, its precision is substantially lower than the other methods. This suggests that Wikipedia is a great source of a vast amount of relevant information, but the other search methods and more diverse types of documents should not be ignored if one needs more precise and nuanced counterevidence.

\begin{table}[t]
    \centering
    \small
    \begin{tabularx}{\linewidth}{l XXX p{2mm} XXX} \toprule
         & \multicolumn{3}{c}{CMV} & & \multicolumn{3}{c}{Kialo} \\
         \cmidrule{2-4} \cmidrule{6-8}
         & P & R & F & & P & R & F \\
        \midrule
        Wikipedia & 42.4 & 64.5 & 51.1 & & 55.1 & 60.8 & 57.8 \\
        Bing & 53.1 & 66.2 & 58.9 & & 59.5 & 63.5 & 61.4 \\
        Google & 47.0 & 64.8 & 54.5 & & 59.9 & 62.6 & 61.2 \\
        \bottomrule
    \end{tabularx}
    \caption{Accuracy of counterevidence retrieval by search methods.}
    \label{tab:evidence_accs_search}
\end{table}

\subsection{Utility of Document Types\label{app:utility_document_types}}
One question we want to answer is: what types of documents are useful sources of counterevidence to argumentative statements? Prior work focuses mostly on Wikipedia articles \cite{thorne-etal-2018-fever,Hua:2018.gen}, debates \cite{Orbach.2020.echo,Wachsmuth:2018retr}, and occasionally news articles \cite{Hua:2019.gen}. In contrast, our candidate sentences come from more diverse types of documents, such as academic papers and government reports. To analyze the utility of different document types, we first annotated each candidate sentence with 13 different types using MTurk (Table~\ref{tab:evidence_domains}). See Appendix~\ref{sec:annot_cand_types} for annotation details.

\begin{figure*}[t]
    \centering
    \includegraphics[width=.8\linewidth]{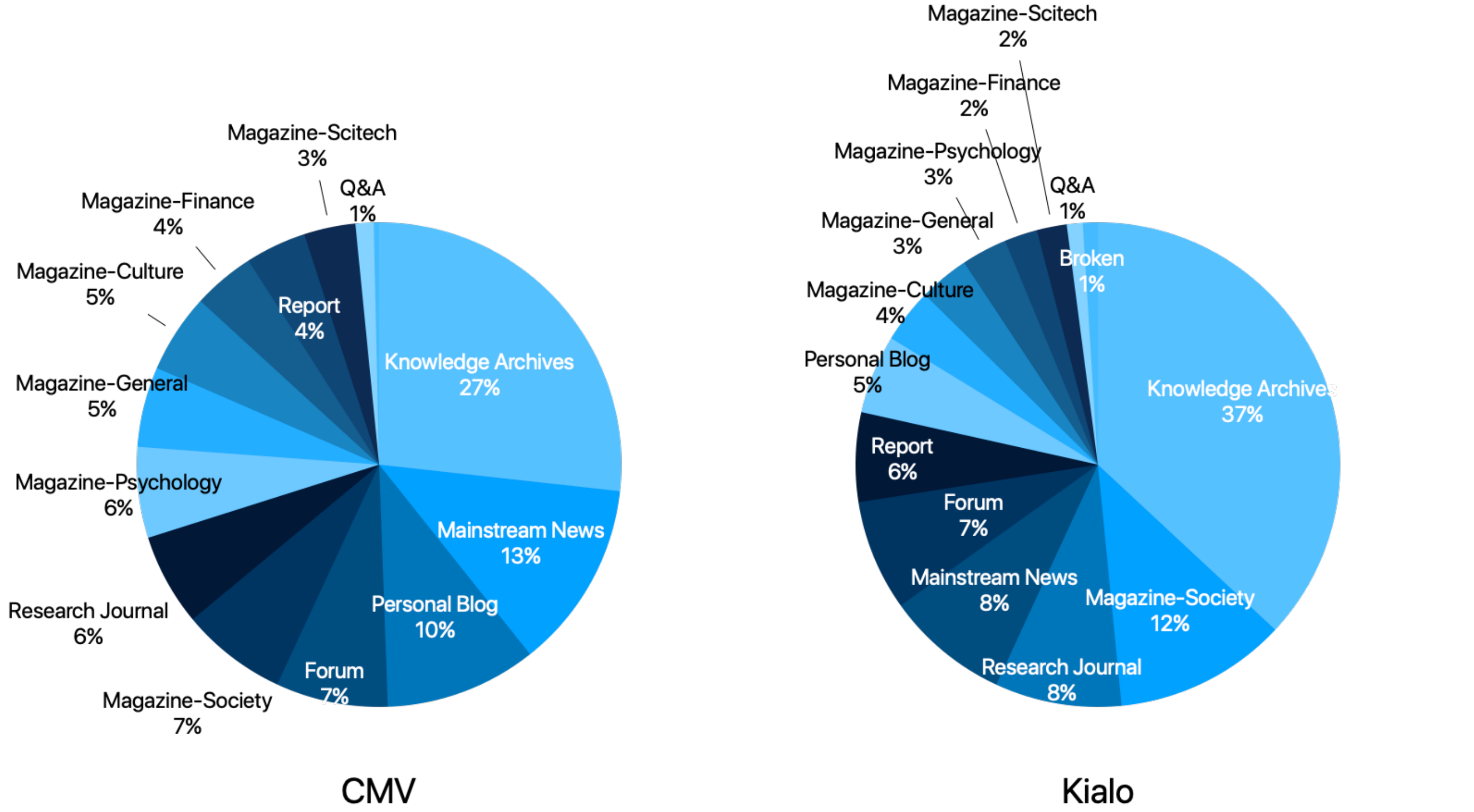}
    \caption{Distribution of document types for valid counterevidence.}
    \label{fig:evidence_domain_dist}
\end{figure*}

\begin{figure*}[t]
    \centering
    \includegraphics[width=.95\linewidth]{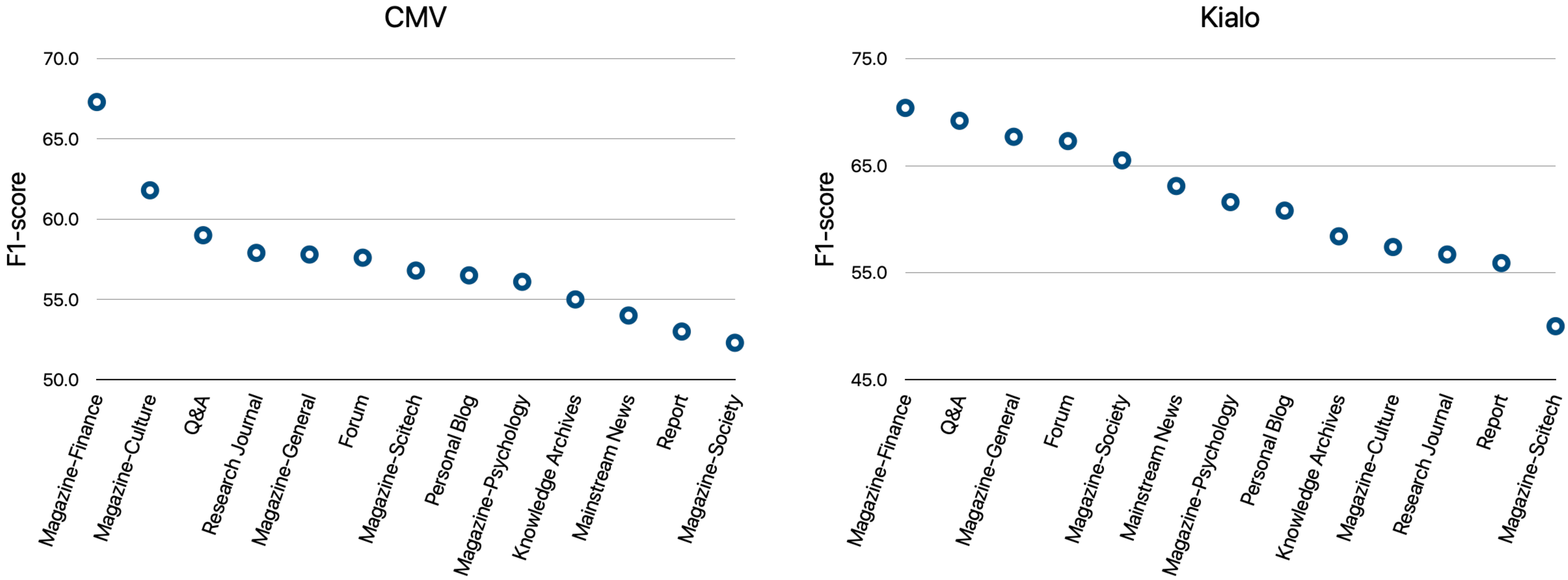}
    \caption{F1-scores of counterevidence retrieval by evidence document types.}
    \label{fig:evidence_domain_accs}
\end{figure*}

First of all, Figure~\ref{fig:evidence_domain_dist} shows the distribution of document types for valid counterevidence. A lot of counterevidence exists in knowledge archives (27--37\%), followed by mainstream news (8--13\%),  magazines about social issues (7--12\%), personal blogs (5--10\%), and research journals (6--8\%). This suggests the benefit of using broader types of documents in counterevidence and fact verification than conventionally used Wikipedia and debates. 

Table~\ref{fig:evidence_domain_accs} summarizes the F1-score of counterevidence retrieval by document types (averaged across all models). For both CMV and Kialo, financial magazines and Q\&A platforms are useful document types providing high F1-scores. For CMV, magazines about culture and research journals are beneficial, while in Kialo, general-domain magazines and forums are useful types. As we also observed in the earlier analysis of search methods, Wikipedia, which is conventionally used in fact verification, and mainstream news are relatively less reliable. So are reports that contain a lot of detailed information.

\subsection{Attackability\label{app:attackability}}
\begin{figure}[t]
    \centering
    \includegraphics[width=.95\linewidth]{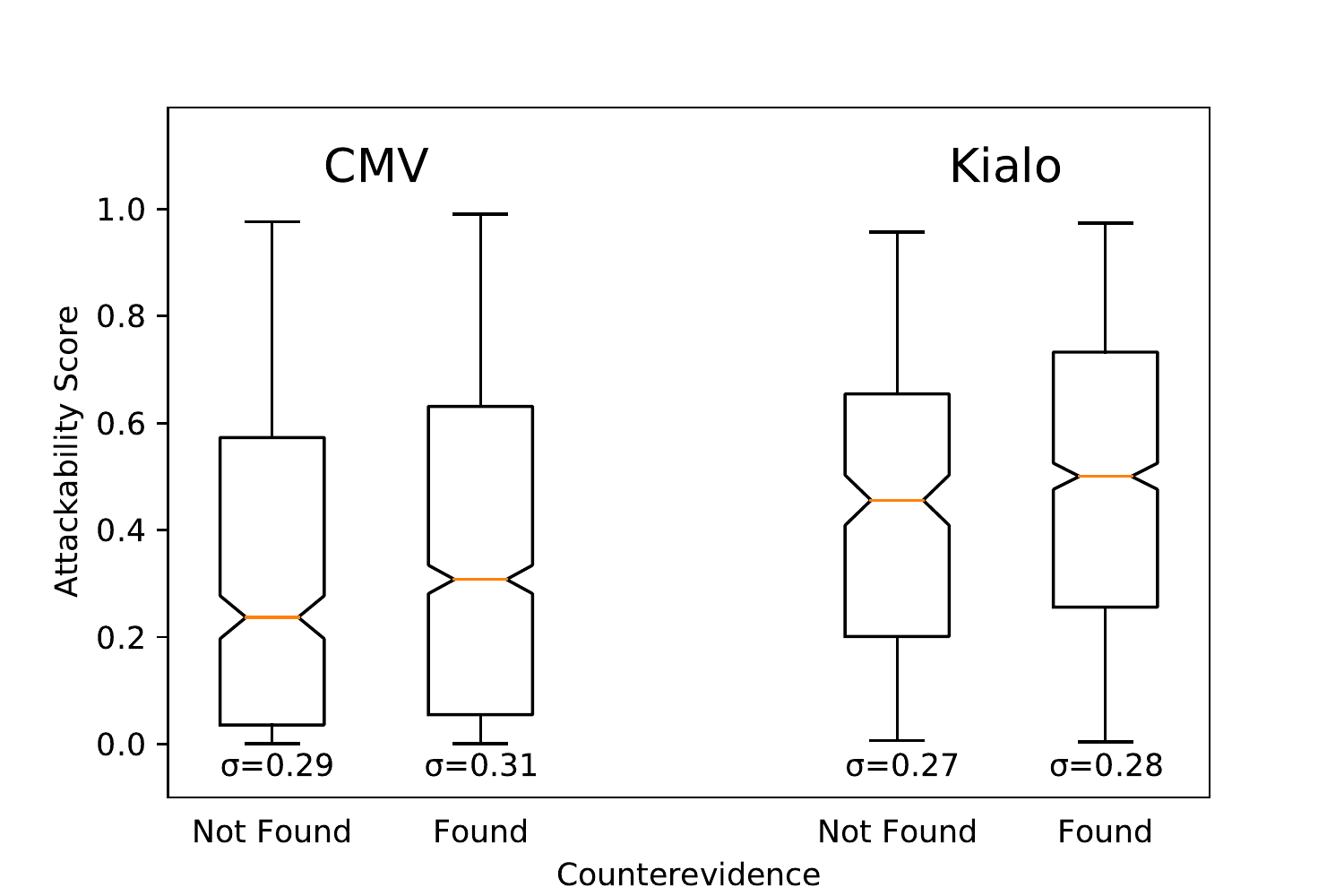}
    \caption{Attackability.}
    \label{fig:attack}
\end{figure}

Our system was originally designed in consideration of the scenario where we counter an argument by first detecting attackable sentences and then finding proper counterevidence to them. Detecting attackable sentences in arguments has recently been studied for CMV based on persuasion outcomes \cite{Jo2020:attackable}. Here, we test if this method can help us find statements for which counterevidence exists. 

We assume that statements in our dataset are attackable if they have at least one candidate sentence that is valid counterevidence.
Figure~\ref{fig:attack} shows the distribution of the attackability scores of statements for which counterevidence was found (Found) and statements for which no counterevidence was found (Not Found). As expected, the attackability scores of statements that have counterevidence are higher than the other statements for both CMV ($p=0.001$) and Kialo ($p=0.003$ by the Wilcoxon rank-sum test). 

The attackability score of each statement also has a small positive correlation with the number of candidate sentences that are valid counterevidence, resulting in Kendall's $\tau$ of 0.057 (CMV, $p=0.002$) and 0.085 (Kialo, $p=0.006$). These results suggest that synergy can be made by integrating our system with attackability detection to build a complete counterargument generation system. We leave this direction to future work.

% \section{Comparison between Bing and Google}
% \begin{table}[t]
%     \centering
%     \small
%     \begin{tabularx}{\linewidth}{p{1cm}XX} \toprule
%          & Bing & Google \\
%         \midrule
%         Top & quizlet.com & ncbi.nlm.nih.gov \\
%         Netlocs & psychologytoday.com & nytimes.com \\
%          & researchgate.net & washingtonpost.com \\
%          & reddit.com & bbc.com \\
%          & quora.com & pewresearch.org \\
%         \midrule
%         Categories &  &  \\
%         of Top &  &  \\
%         Sites &  &  \\
%          &  &  \\
%         \midrule
%         Top & com (68\%) org (16\%) &  com (45\%) org (25\%) \\
%         Domains & edu (4\%) net (4\%) & gov (11\%) edu (10\%) \\
%         \midrule
%         Types & pdf (1.5\%) txt (1.2\%) & pdf (12.3\%) txt (0.2\%) \\
%         \bottomrule
%     \end{tabularx}
%     \caption{Caption}
% \end{table}

\section{Reproducibility Checklist}
\subsection{Knowledge-Enhanced NLI\label{app:reproducibility_kenli}}
\begin{itemize}[topsep=0pt,itemsep=0pt,parsep=0pt,partopsep=0pt,leftmargin=10pt]
    \item \textbf{A clear description of the mathematical setting, algorithm, and/or model:} Explained in the main text.
    \item \textbf{Submission of a zip file containing source code, with specification of all dependencies, including external libraries, or a link to such resources:} This information will be made available at a git repository upon publication. 
    \item \textbf{Description of computing infrastructure used:} Intel(R) Xeon(R) Gold 5215 CPU @ 2.50GHz (20 CPUs), 128GiB System memory, Quadro RTX 8000 (4 GPUs).
    \item \textbf{The average runtime for each model or algorithm (training + inference):} KENLI+C: 311.9 mins / KENLI+E: 562.8 mins
    \item \textbf{Number of parameters in each model:} KENLI+C: 135,676,421 / KENLI+E: 174,330,206.
    \item \textbf{Corresponding validation F1-score (across all datasets) for each reported test result:} KENLI+C: 74.5 / KENLI+E: 75.0.
    \item \textbf{Explanation of evaluation metrics used, with links to code:} Explained in the main text.

    \item \textbf{The exact number of training and evaluation runs:} 5 runs.
    \item \textbf{Bounds for each hyperparameter:} None.
    \item \textbf{Number of hyperparameter search trials:} No hyperparameter search.
    \item \textbf{Hyperparameter configurations for best-performing models:} Explained in the main text.

    \item \textbf{Relevant details such as languages, and number of examples and label distributions:} Explained in the main text.
    \item \textbf{Details of train/validation/test splits:} Explained in the main text.
    \item \textbf{Explanation of any data that were excluded, and all pre-processing steps:} Explained in the main text.
    \item \textbf{A zip file containing data or link to a downloadable version of the data:} Example-NLI will be made available in a git repository upon publication.
    \item \textbf{For new data collected, a complete description of the data collection process, such as instructions to annotators and methods for quality control:} Explained in the main text.
\end{itemize}

\subsection{Evidence Retrieval\label{app:reproducibility_retr}}
\begin{itemize}[topsep=0pt,itemsep=0pt,parsep=0pt,partopsep=0pt,leftmargin=10pt]
    \item \textbf{A clear description of the mathematical setting, algorithm, and/or model:} Explained in the main text.
    \item \textbf{Submission of a zip file containing source code, with specification of all dependencies, including external libraries, or a link to such resources:} This information will be made available at a git repository upon publication.
    \item \textbf{Description of computing infrastructure used:} Intel(R) Xeon(R) Gold 5215 CPU @ 2.50GHz (20 CPUs), 128GiB System memory, Quadro RTX 8000 (4 GPUs).
    \item \textbf{The average runtime for each model:} 
        \begin{itemize}
            \item Document retrieval: 36.1 mins.
            \item Relation classification: KENLI+C: 44.3 mins / KENLI+E: 71.2 mins.
        \end{itemize}
    \item \textbf{Number of parameters in each model:}
        \begin{itemize}
            \item Document retrieval: 125,853,255.
            \item Relation classification: KENLI+C: 135,676,421 / KENLI+E: 174,330,206.
        \end{itemize}
    \item \textbf{Corresponding validation performance for each reported test result:} No validation.
    \item \textbf{Explanation of evaluation metrics used, with links to code:} Explained in the main text.

    \item \textbf{The exact number of evaluation runs:} 1 run.
    \item \textbf{Bounds for each hyperparameter:} None.
    \item \textbf{Number of hyperparameter search trials:} No hyperparameter search.

    \item \textbf{Relevant details such as languages, and number of examples and label distributions:} Explained in the main text.
    \item \textbf{Details of train/validation/test splits:} Explained in the main text.
    \item \textbf{Explanation of any data that were excluded, and all pre-processing steps:} Explained in the main text.
    \item \textbf{A zip file containing data or link to a downloadable version of the data:} The annotated data will be made available at a git repository upon publication.
    \item \textbf{For new data collected, a complete description of the data collection process, such as instructions to annotators and methods for quality control:} Explained in the main text.
\end{itemize}